\documentclass{article}

\usepackage[preprint]{neurips_2024}

\usepackage[utf8]{inputenc} \usepackage[T1]{fontenc}    \usepackage{hyperref}       \usepackage{url}            \usepackage{booktabs}       \usepackage{amsfonts}       \usepackage{nicefrac}       \usepackage{microtype}      \usepackage{xcolor}

\usepackage[normalem]{ulem}                \usepackage[noabbrev,capitalize]{cleveref} \usepackage{subcaption}                                \usepackage{comment}                       \usepackage{wrapfig}

\usepackage{longtable}
\usepackage{relsize}                       \usepackage{multirow}
\usepackage{adjustbox}

\usepackage{pifont}
\newcommand{\cmark}{\ding{51}}
\definecolor{light-gray}{gray}{0.75}
\newcommand{\xmark}{\color{light-gray}\ding{55}}

\usepackage{array}
\newcolumntype{R}[2]{>{\adjustbox{angle=#1,lap=\width-(#2)}\bgroup}l<{\egroup}}
\newcommand*\rot{\multicolumn{1}{R{45}{1em}}}\newcommand{\fullrot}[1]{\rotatebox[origin=l]{90}{#1}}
\newcommand{\norot}[1]{\rotatebox[origin=l]{0}{#1}}

\title{Towards Foundation Models\\for Critical Care Time Series}

\author{Manuel Burger\thanks{equal contribution}\:\:\thanks{Department of Computer Science, ETH Zurich, Switzerland}\\
    \texttt{manuel.burger@inf.ethz.ch}
    \And
    Fedor Sergeev\footnotemark[1]\:\:\footnotemark[2]\\
    \texttt{fedor.sergeev@inf.ethz.ch}
    \AND
    Malte Londschien\thanks{Seminar for Statistics, ETH Zürich, Switzerland}\:\:\thanks{AI Center, ETH Zurich, Switzerland}\:\:\footnotemark[2]
    \And
    Daphné Chopard\footnotemark[2]\:\:\thanks{Department of Intensive Care and Neonatology and Children's Research Center, University Children's Hospital Zurich, University of Zurich, Switzerland}
    \And
    Hugo Yèche\footnotemark[2]
    \And
    Eike Gerdes\thanks{University of Zurich, Switzerland}
    \And
    Polina Leshetkina\thanks{Department of Health Science and Medicine, University of Luzern, Switzerland}
    \And
    Alexander Morgenroth\footnotemark[2]
    \And
    Zeynep Babür\thanks{Zurich University of Applied Sciences (ZHAW), Switzerland}
    \And
    Jasmina Bogojeska\footnotemark[8]
    \And
    Martin Faltys\thanks{Department of Intensive Care Medicine, University Hospital and University of Bern, Switzerland}
    \AND
    Rita Kuznetsova\footnotemark[1]\:\:\footnotemark[2]\\
    \texttt{rita.kuznetsova@inf.ethz.ch}
    \And
    Gunnar Rätsch\footnotemark[1]\:\:\footnotemark[2]\\
    \texttt{raetsch@inf.ethz.ch}
}

\begin{document}

\maketitle

\begin{abstract}
    Notable progress has been made in generalist medical large language models across various healthcare areas. However, large-scale modeling of in-hospital time series data - such as vital signs, lab results, and treatments in critical care - remains underexplored. Existing datasets are relatively small, but combining them can enhance patient diversity and improve model robustness. 
    To effectively utilize these combined datasets for large-scale modeling, it is essential to address the distribution shifts caused by varying treatment policies, necessitating the harmonization of treatment variables across the different datasets. 
    This work aims to establish a foundation for training large-scale multi-variate time series models on critical care data and to provide a benchmark for machine learning models in transfer learning across hospitals to study and address distribution shift challenges. We introduce a harmonized dataset for sequence modeling and transfer learning research, representing the first large-scale collection to include core treatment variables. Future plans involve expanding this dataset to support further advancements in transfer learning and the development of scalable, generalizable models for critical healthcare applications.
\end{abstract}

\section{Introduction}\label{intro}

Foundation models trained on complex multi-modal medical data have the potential to significantly transform healthcare~\citep{moor2023foundation}. Considerable advancements have been made in the development of generalist medical Large Language Models (LLM)~\citep{singhal2023large, chen2023meditron70b}, computer vision models in pathology~\citep{vorontsov2024virchowmillionslidedigitalpathology}, single-cell multi-omics models~\citep{cui2024scGPT}, and sequence models on coded Electronic Health Records (EHR)~\citep{wornow2023ehrshotehrbenchmarkfewshot}. 

One area that remains underexplored is the foundation models for critical care time series\footnote{We call \emph{critical care} a setting, where a patient is being closely monitored (e.g., in emergency departments and intensive care units, during surgery, etc.)}. It is promising because of the prospective benefits for patients and the availability of large (multi-site and multi-national) and rich (multi-variate, including vital signs, lab measurements, and treatments) data.

Developing a large-scale foundation model with robust generalization capabilities across hospitals and countries requires a comprehensive dataset with high patient diversity~\citep{rockenschaub2024impact}. Individually published datasets from Intensive Care Units (ICU) and Emergency Departments (ED)~\citep{johnson2016mimic, hirid, ams, rodemund2023salzburg, johnson_mimic-iv-ed_2023, wornow2023ehrshot} are relatively small compared to modern standards in fields such as Natural Language Processing (NLP)~\citep{gao2020pile800gbdatasetdiverse}. However, by aggregating them, it is possible to scale the number of admissions by an order of magnitude and increase their diversity. Previous works addressing such aggregation did not include all the ICU datasets, add ED datasets, or harmonize treatment variables \citep{bennett2023ricu, yang2023pyhealth}.

A key challenge in creating a foundation model is to ensure its robustness to distribution shifts. In the clinical domain, this is especially difficult because of substantial differences in recording formats and treatment policies between hospitals and countries~\citep{huser2024medrxiv_respiratory}. Robustness to these shifts would suggest that the model generalizes beyond cohort-specific pattern matching and achieves a deeper understanding of human physiology. Specifically on critical care time series, most previous works considered single-center performance~\citep{harutyunyan2019multitask, yeche2021, chen2024multimodal}. The few publications that did consider transfer, either focused on a specific task \cite{moor2023predicting} or did not attempt to improve model generalization and ensure its robustness~\citep{van2023yet}.

Our aim is to establish the foundation for training and evaluating large-scale multi-variate time-series models on real-world hospital data from critical care. To achieve this goal, we create a large multi-center dataset covering a wide array of clinical features and build an understanding of what machine learning algorithms work well on such data. 

We expect this work to become the basis for a future foundational model with a wide range of downstream medical applications. Specifically, it will unlock research for small cohorts of specific patients using few-shot learning or fine-tuning, mirroring the impact of pretrained language models in NLP. Furthermore, for the ML community, the dataset we present will be a valuable resource for research into sequence modeling, meta and transfer learning, domain adaptation, and generalization.

Our current contributions are two-fold:
\begin{itemize}
    \item \emph{Dataset}. We introduce the largest harmonized critical care time series medical dataset. It is the first of such datasets to (a) harmonize the core treatment variables, (b) include datasets from both ICU and ED, (c) incorporate data from Asia in addition to Europe and the USA, and (d) provide annotations and results on multiple organ failure tasks on the same data. It is extendable and can be used for research in sequence modeling, domain generalization, and meta-learning.

    \item \emph{Benchmark}. We run a comprehensive benchmark of machine learning models on the new dataset. We perform transfer studies and evaluate performance on clinically relevant real-time prediction tasks in-distribution as well as out-of-distribution.

\end{itemize}
 \section{Related Work}\label{rel_work}

\paragraph{Time Series Foundation Models} Following advances in natural language processing (NLP), Large-scale multi-purpose pretrained models referred to as "foundation models" have sprouted across data and application types. When considering time series, a large body of works has exclusively focused on forecasting non-medical data~\citep{das2023decoder, goswami2024moment, zhou2023one, ansari2024chronos}.

\paragraph{Foundation Models for Healthcare}
In the clinical domain, existing foundation models have not considered in-hospital critical care time series but rather other forms of Electronic Health Records (EHR) such as billing codes~\citep{wornow2023ehrshotehrbenchmarkfewshot} and medical reports~\citep{chen2023meditron70b}. Some studies~\citep{guo2024multi, wornow2023shaky} explored the adaptability of public EHR models for clinical prediction tasks, while others~\citep{guo2023ehr} evaluated their effectiveness in improving in-distribution and out-of-distribution performance.

\paragraph{Benchmarks on ICU time-series} 
In the literature, we observe two benchmarking strategies: single- and multi-center.  \citet{harutyunyan2019multitask} provided the first standardized and reproducible single-center benchmarks built on MIMIC-III~\citep{MIMIC-III}. Following this seminal work, a line of studies emerged that defined new tasks or explored different datasets, such as \citet{sheikhalishahi2020benchmarking} on eICU~\citep{pollard2018eicu},~\citet{yeche2021} on HiRID~\citep{hirid}, and~\citet{wang2020mimicextract} as an alternative on MIMIC-III. The proliferation of work around single-center data has led researchers to aggregate them into multi-center studies such as~\cite{moor2023predicting} and~\cite{van2023yet}. We present a comparison in~\cref{tab:related_works_small}. It is important to emphasize that, unlike previous efforts, we both perform a new largest to-date dataset harmonization and build a comprehensive benchmark.

\begin{table}[tb]
\centering
\resizebox{0.9\textwidth}{!}
{\begin{tabular}{l *{15}{l}}
    \toprule
    & \multicolumn{9}{c}{Datasets} & \multicolumn{2}{c}{Features} & \multicolumn{3}{c}{Transfer}  \\ \cmidrule(lr){2-10} \cmidrule(rl){11-12} \cmidrule(lr){13-15} 
     & \fullrot{MIMIC-III} & \fullrot{eICU} & \fullrot{UMCdb} & \fullrot{HiRID} & \fullrot{MIMIC-IV} & \fullrot{SICdb} & \fullrot{PICdb} & \fullrot{Zigong EHR} & \fullrot{MIMIC-IV-ED} & \fullrot{Harm.\ treatments} & \fullrot{Multi-unit} & \fullrot{Single-center} & \fullrot{Multi-center} & \fullrot{Fine-tuning}  \\
    \midrule
    \cite{tang2020democratizing} & \cmark & \cmark & \xmark & \xmark & \xmark & \xmark & \xmark & \xmark & \xmark & \xmark & \xmark & \xmark & \xmark & \xmark  \\
    \cite{moor2023predicting} & \cmark & \cmark & \cmark & \cmark & \xmark & \xmark & \xmark & \xmark & \xmark &  \xmark & \xmark & \xmark & \cmark & \xmark  \\
    \cite{van2023yet} & \cmark & \cmark & \cmark & \cmark & \cmark & \xmark & \xmark & \xmark & \xmark & \xmark & \xmark & \cmark & \cmark & \xmark  \\ 
    \addlinespace
    Ours & \cmark & \cmark & \cmark & \cmark & \cmark & \cmark & \cmark & \cmark & \cmark & \cmark & \cmark & \cmark & \cmark & \cmark \\
    \bottomrule
\end{tabular}}
\caption{Multi-dataset critical care time-series benchmarks}
\label{tab:related_works_small}
\end{table}

  \section{Data Harmonization and Processing}\label{benchmark}

\subsection{Data Sources}

To maximize the number of harmonized physiological measurements and treatment data points, we incorporate all ICU datasets that are freely available to the academic community. These include datasets from the USA (MIMIC-III~\citep{MIMIC-III}, MIMIC-IV~\citep{johnson2023mimic, MIMIC-III}, and eICU~\citep{pollard2018eicu}), Europe (AmsterdamUMCdb~\citep{ams}, SICdb~\citep{rodemund2023salzburg}, and HiRID~\citep{hirid}), and China  (PICdb~\citep{li2019paediatric} and Zigong EHR~\citep{zigong-ehr-dataset}). Additionally, we incorporate an ED dataset~\citep{johnson_mimic-iv-ed_2023}. Most datasets are available on the Physionet Platform~\citep{goldberger2000physiobank} or directly with the dataset provider (e.g. AmsterdamUMCdb~\citep{ams}). The dataset overview and statistics are shown in~\cref{tab:datasets_stats}, Appendix~\ref{sec:app_data}.

To the best of our knowledge, this is the first work bringing together critical care datasets from the US, Europe, and, for the first time, China. Harmonizing datasets across different continents can improve generalization and is crucial to the fairness and inclusiveness of ML research on critical care data. 
The diversity of our dataset enables research for small but specific cohorts of patients. For example, PICdb~\citep{li2019paediatric} is a small pediatric dataset. The average age is under one year, while it is over 60 on other datasets (see  \cref{tab:datasets_stats} in Appendix~\ref{sec:app_data}). By providing an easy way to pretrain on large amounts of data, we create an opportunity for smaller-scale targeted studies to benefit from the existing larger-scale research on modeling for critical care time series.

By incorporating both ICU and ED datasets, we provide a way to study joint ED-ICU models, potentially leading to a unified clinical prediction model regardless of the hospital unit.

\subsection{Data Harmonization}  

The datasets we consider are recorded using different, non-standardized, formats. We perform dataset harmonization with the \texttt{ricu} package as a basis \citet{bennett2023ricu}. \texttt{ricu} defines data source agnostic concepts as an abstraction for encoding clinical concepts. These include static information about the patient (e.g., height), observations (e.g., heart rate), and treatments (e.g., administration of antibiotics). By mapping the concepts to source variables from each dataset the package facilitates exporting a unified view of the data across all of them.

Expanding prior work~\citep{van2023yet, moor2023foundation, bennett2023ricu}, we implement new observation concepts and incorporate new ICU datasets, namely SICdb, PICdb, Zigong EHR, creating the largest harmonized ICU dataset to date. Further, we integrate an ED dataset (MIMIC-IV-ED), increasing the number of processed stays from around 400,000 \citep{van2023yet} to over 600,000. The expansion increases the total number of final extracted individual data points from approximately 400 million close to one billion.

Crucially, we introduce a principled way to harmonize a wide range of treatment variables. This significantly increases the number of concepts compared to previous works~\citep{moor2023foundation, van2023yet}. We define the new concepts using clinical expert opinion informed by what variables were reported as most important for various tasks and models in the literature (see \cref{tab:metavariables_literature}, Appendix~\ref{sec:app_data}). Previous works \citep{moor2023foundation} have suggested that including medication variables harms the accuracy of predictive models, but little research has been done into the reasons behind this effect and what can be done to mitigate it. The information about administered medications is an insight into the actions of the clinicians, and could drastically improve the model accuracy and transfer. By including these variables in the harmonization pipeline, we prepared the ground for deeper investigation in this direction.

\citet{blendedicuOliver2023} have proposed a processing pipeline for a subset of the source datasets considered in this work and harmonized treatments by including drug exposure information as indicators. We improve on this by (1) considering not only indicators, but also administration rates for core medications used in critical care settings, and (2) grouping individual drugs into abstract treatment concepts, thereby increasing the overlap across datasets in concepts while maintaining relevance for downstream applications.

Ultimately, providing harmonized treatment information including administered dosages across a collection of datasets enables future research on learning generalizable treatment effect estimations on critical care time series.

\subsection{Processing pipeline}  

We use anonymized data with permissive exclusion criteria (\cref{sec:inclusion-criteria}) to include as many patients as possible. The time series are extracted as a uniform grid at resolutions of 5 and 60 minutes depending on the dataset balancing sampling precision and interoperability (see \cref{tab:datasets_stats}, Appendix~\ref{sec:app_data}). Further, similar to~\citet{yeche2021}, we remove outliers, impute missing values, scale variables, extract features for tree-based models, and define task labels.

Finally, we export the processed data into two formats consumable by modern deep learning and classical machine learning algorithms. First, a dense fully imputed time-grid (including feature extraction if applicable for the model) and second a tokenized data format~\citep{gorishniy2023revisitingdeeplearning, horn2020setfunctionstimeseries}, which encodes only ground truth measured data points as a triplet of time, variable, and observed value. The second format removes the need for imputation and has recently been proposed as a more suitable data representation format for scaling models on highly irregular time-series data~\citep{tipirneni2022selfsupervisedtransformer} and sharing of data processing outputs~\citep{arnrichmedical}. Further details on data harmonization and processing are described in \cref{sec:app_data}.

\begin{wrapfigure}{r}{0.5\textwidth}
\centering
    \includegraphics[width=\linewidth]{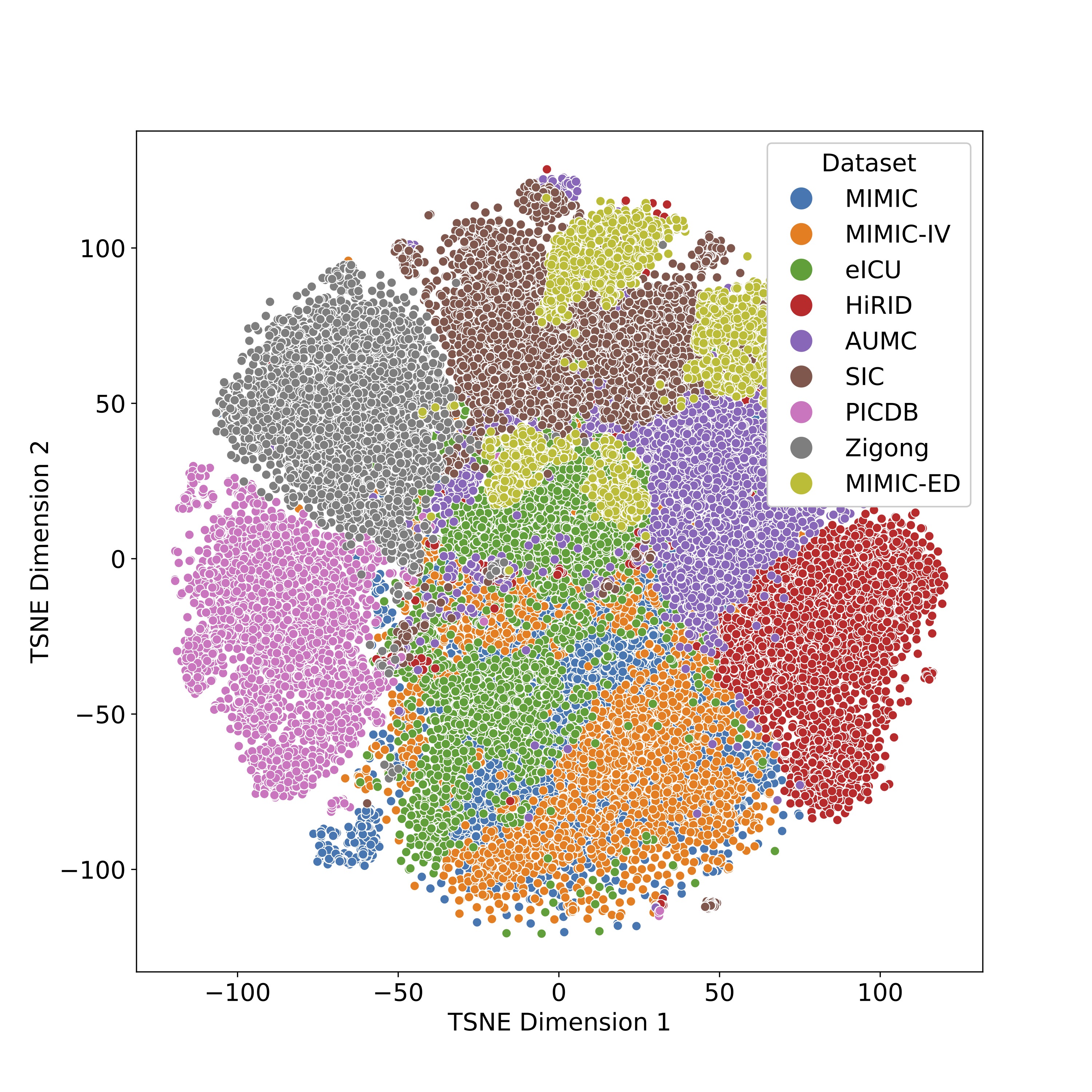}
    \caption{Visualization of harmonized and processed data by t-SNE~\citep{vandermaaten08a_tsne}. Each point represents a time step.}
    \label{fig:tsne-data} 
\end{wrapfigure}

\cref{fig:tsne-data} shows visualization of harmonized and processed data by t-SNE~\citep{vandermaaten08a_tsne}. The apparent clustering by source hospital emphasizes the challenge of developing a predictor that is robust to these distribution shifts across sites.

\subsection{Task Annotations}

Our study focuses on clinically relevant real-time prediction tasks where patient outcomes in the ICU can be influenced by timely intervention. These include: circulatory failure~\citep{hyland2020early}, respiratory failure~\citep{huser2024medrxiv_respiratory}, and kidney function~\citep{lyu2024empirical}. Additionally, to ensure a diverse range of tasks and to facilitate comparison with previous works, we include the prediction of decompensation~\citep{harutyunyan2019multitask}. All of these are modeled as binary early event prediction tasks~\citep{yeche2024dsaforeep} (i.e. forecasting and prognosing future patient states) with a clinically relevant prediction horizon. On the emergency department data, we consider disposition prediction~\citep{chen2024multimodal, lee2020prediction}. 

 \section{Experiments}\label{exp}

\subsection{Setup}

\paragraph{Models} 

In our proposed benchmark, we considered two groups of machine learning algorithms. The first group consists of classical machine learning methods (LightGBM~\citep{ke2017lightgbm} for gradient boosted decision trees and regularized Linear Regression~\citep{glum2020quantco}), which are highly effective for real-time prediction tasks on critical care time series~\citep{harutyunyan2019multitask,hyland2020early,yeche2021}. For these models, we either use the forward-filled last available measurement for each variable (\emph{Last Meas.)} or include hand-extracted features from the history based on the work by~\citet{soenksen2022multimodalai}, which we further expanded to improve performance~(\cref{sec:app_features}). The second group is focused on deep learning methods. We select established and state-of-the-art sequence architectures for this group: Gated Recurrent Unit (GRU)~\citep{cho2014learning}, Transformer~\cite{vaswani2017attention}, Mamba~\citep{dao2024transformersssmsgeneralizedmodels} and xLSTM~\citep{beck2024xlstm}.

\paragraph{Training} Deep learning approaches were implemented in \texttt{pytorch}~\citep{paszke2019pytorch} and trained using AdamW optimizer~\citep{loshchilov2019decoupledweightdecayregularization}, with a cross-entropy objective for classification tasks. We evaluated models using task-specific metrics: the area under the receiver operating characteristic curve (AUROC) and the area under the precision-recall curve (AUPRC) for classification tasks. All metrics were computed using \texttt{torchmetrics}~\citep{torchmetrics}. For all models, we tuned a subset of important hyper-parameters using grid search. Each set of parameters was run with 3 different random initializations and we report mean metric performance (standard deviations are shown in tables if space permits it).

\subsection{In-distribution Benchmark}

\begin{table*}[tb]
\centering
\resizebox{\textwidth}{!}{\begin{tabular}{l l c c c c c c c c c c c c c c}
    \toprule
    \bf Dataset & \bf Task & \multicolumn{7}{c}{\textbf{Single-Center}} & \multicolumn{7}{c}{\textbf{Multi-Center}} \\
    \cmidrule(lr){3-9} \cmidrule(lr){10-16} 
     & & \rot{LR (Last Meas.)} & \rot{LGBM (Last Meas.)} & \rot{LGBM (Feat.)} & \rot{GRU} & \rot{Transformer} & \rot{Mamba}  & \rot{xLSTM} & \rot{LR (Last Meas.)} & \rot{LGBM (Last Meas.)} & \rot{LGBM (Feat.)} & \rot{GRU} & \rot{Transformer} & \rot{Mamba}  & \rot{xLSTM} \\

\midrule
    \multirow{4}{*}{\norot{MIMIC-IV}} & Dec. 24h & 93.7 & 95.4 & \textbf{97.3} & 95.8 & 95.7 & 95.8 & 95.8 & 92.0 & 94.7 & 96.1 & 95.8 & 95.0 & 95.0 & 95.7 \\
    & Circ. 8h  & 93.5 & 95.2 & \textbf{95.6} & 94.9 & 94.9 & 94.8 & 95.0 & 92.5 & 94.6 & 95.1 & 94.5 & 94.4 & 94.5 & 94.7 \\
    & Resp. 24h & 76.4 & 79.7 & \textbf{81.1} & 79.9 & 79.7 & 79.7 & 79.8 & 74.1 &     78.7  &   79.9 &  79.3  &  78.9  &   79.5 &   79.5   \\
    & Kidn. 48h & 83.2 & 87.5 & \textbf{89.8} & 88.3 & 87.7 & 87.9 & 88.3 & 81.6 &     86.9  &  89.2  &  87.8  &  86.2  &   88.1 &   88.3  \\

\midrule
    \multirow{4}{*}{\norot{eICU}} & Dec. 24h & 91.1 & 93.0 & \textbf{95.8} & 93.4 & 93.3 & 93.4 & 93.5 & 89.1 &     92.4  &          93.9 &  92.8  &  92.4  &   92.9 &   93.1 \\
    & Circ. 8h  & 94.4 & 95.6 & \textbf{96.0} & 95.4 & 95.4 & 95.2 & 95.3   &    93.2 &     95.2  &          95.6 &  94.7  &  94.8  &   94.8 &   94.9 \\
    & Resp. 24h & 79.2 & 82.1 & \textbf{83.6} & 82.8 & 82.3 & 82.2 & 82.1 & 78.1 &     81.7  &          82.5 &  81.8  &  81.3  &   81.8 &   82.1 \\
    & Kidn. 48h & 74.5 & 82.0 & \textbf{85.6} & 83.7 & 82.9 & 83.3 & 83.8 & 73.0 &     81.0  &          84.2 &  82.1  &  81.5  &   82.7 &   83.1 \\

\midrule
    \multirow{4}{*}{\norot{HiRID}} & Dec. 24h & 93.0 & 93.8 & 94.5 & \textbf{94.6} & 94.0 & 94.4 & 94.3 & 92.4 &     94.4  &          95.1 &  94.3  &  94.1  &   94.4 &   94.5 \\
    & Circ. 8h  & 90.8 & 91.9 & \textbf{92.6} & 92.3 & 92.0 & 92.0 & 92.2 & 90.7 &     92.1  &          92.8 &  \textbf{92.6}  &  92.4  &   92.3 &   \textbf{92.6} \\
    & Resp. 24h & 75.2 & 76.6 & 78.1 & 77.2 & 76.9 & 76.6 & 76.8 & 75.1 &     77.1  &          \textbf{78.4} &  78.0  &  77.5  &   77.4 &   77.3 \\
    & Kidn. 48h & 91.2 & 93.0 & 93.7 & 92.3 & 91.1 & 91.9 & 92.1 & 90.7 &     93.4  &          \textbf{94.3} &  93.6  &  93.2  &   93.0 &   93.4 \\

\midrule
    \multirow{4}{*}{\norot{UMCdb}} & Dec. 24h & 88.9 & 92.3 & \textbf{95.9} & 92.9 & 91.8 & 92.1 & 92.3 & 88.4 &     92.3  &          95.2 &  93.2  &  92.9  &   93.4 &   93.4 \\
    & Circ. 8h  & 96.2 & 97.1 & 97.5 & 97.6 & 97.3 & 97.3 & 97.5 & 95.7 &     97.0  &          97.7 &  \textbf{97.8}  &  97.7  &   97.7 &   \textbf{97.8} \\
    & Resp. 24h & 78.8 & 80.4 & \textbf{82.0} & 81.2 & 80.7 & 80.6 & 80.5 & 78.2 &     80.5  &          82.1 &  81.4  &  81.2  &   80.8 &   80.7 \\
    & Kidn. 48h & 93.4 & 95.1 & \textbf{95.6} & 94.5 & 94.1 & 93.9 & 94.3 & 93.4 &     95.8  &          96.1 &  95.2  &  95.0  &   95.2 &   95.0 \\
    
\midrule
    \multirow{4}{*}{\norot{SICdb}} & Dec. 24h & 83.7 & 88.1 & 88.8 & 87.9 & 86.7 & 87.2 & 87.8 & 81.8 &     88.8  &          90.9 &  89.2  &  89.2  &   89.0 &   \textbf{89.6} \\
    & Circ. 8h  & 88.7 & 90.3 & 91.6 & 91.0 & 90.7 & 90.6 & 90.5 & \textbf{95.7} &     90.3  &          91.7 &  91.3  &  91.1  &   91.1 &   91.3 \\
    & Resp. 24h & 77.9 & 80.8 & \textbf{81.4} & 80.7 & 80.2 & 80.3 & 80.1 & 78.2 &     80.9  &          81.7 &  81.1  &  80.7  &   81.2 &   81.0 \\
    & Kidn. 48h & 87.3 & 89.4 & 90.8 & 88.9 & 87.7 & 88.3 & 88.1 & \textbf{93.4} &     90.1  &          91.9 &  89.2  &  89.2  &   89.1 &   89.3 \\

\midrule
    \multirow{4}{*}{\norot{PICdb}} & Dec. 24h & 85.3 & 85.6 & \textbf{90.3} & 87.8 & 88.0 & 87.2 & 87.8 & 70.6 &     87.0  &          88.8 &  83.2  &  81.8  &   84.5 &   84.0 \\
    & Circ. 8h  & 94.2 & 94.7 & \textbf{96.8} & 96.0 & 96.0 & 95.8 & 96.4 &  88.6 &     92.4  &          92.1 &  92.0  &  92.2  &   93.2 &   93.6 \\
    & Resp. 24h & \textbf{71.3} & 66.3 & 68.5 & 68.4 & 70.7 & 70.2 & 65.9 & 65.9 &     59.5  &          59.3 &  66.1  &  66.2  &   66.8 &   67.3 \\
    & Kidn. 48h & 73.5 & \textbf{81.9} & 78.9 & 63.5 & 63.8 & 66.5 & 69.0 &  57.4 &     71.8  &          81.2 &  67.5  &  67.1  &   67.0 &   64.5 \\

\midrule
    \multirow{2}{*}{\norot{Zigong}} & Dec. 24h & 69.3 & 78.3 & \textbf{92.2} & 86.4 & 85.1 & 76.6 & 68.7 & 66.8 &     70.1  &          80.3 &  73.8  &  71.6  &   71.4 &   74.4 \\
    & Circ. 8h  & 88.2 & 92.3 & \textbf{93.5} & 88.8 & 86.6 & 84.8 & 84.7 & 89.0 &     90.0  &          89.1 &  86.0  &  84.9  &   86.0 &   86.2 \\

    \bottomrule
    \end{tabular}
}
    \caption{Benchmarking results in-distribution (AUROC). Bold is best in each row. Multi-center trains on all datasets together and provides the individual test performances. All results show the mean over three different random initialization, except for LR models that are trained using convex optimization. }

    \label{tab:benchmark-id}
\end{table*} 
Single-center in-distribution training represents the classical setting where a model is trained and evaluated on the train and test subsets of a single source dataset.

In multi-center in-distribution setting a model is trained on all the harmonized datasets jointly and evaluated on a test set of a single dataset. By carefully normalizing the features not to depend on resolution and passing appropriate positional encoding at each time step, we can train the model in a multi-resolution fashion. Test results then report the individual performances of this single model on each source dataset separately.

Results for in-distribution training for single- and multi-center training are presented in \cref{tab:benchmark-id}. Results for disposition prediction on MIMIC-IV-ED are presented in \cref{tab:disposition}.

\begin{wraptable}{r}{0.35\textwidth}
\vspace{-0.3cm}
\centering
\resizebox{0.325\textwidth}{!}{
    \centering
    \begin{tabular}{ll}
        \toprule
        Models    & Disposition     \\
        \midrule
        LR (Last Meas.) & $72.7$ \\
        LR      & $78.2$\\
        LGBM (Last Meas.)     &    $74.8 \pm 0.02$   \\
        LGBM (Feat.)  & $\mathbf{80.4 \pm 0.01}$ \\
        GRU   & $79.9 \pm 0.04$ \\
        Transformer     & $79.9 \pm 0.03$ \\
        Mamba   & $79.9 \pm        0.09$ \\
        xLSTM  &$80.1 \pm 0.16$ \\
        \bottomrule
    \end{tabular}
}
\caption{Disposition prediction on MIMIC-IV-ED (AUROC).}
\label{tab:disposition}
\end{wraptable} 

\subsection{Out-of-distribution Benchmark}

\begin{figure}
    \centering
\begin{subfigure}[t]{0.58\textwidth} \centering
        \includegraphics[width=\linewidth]{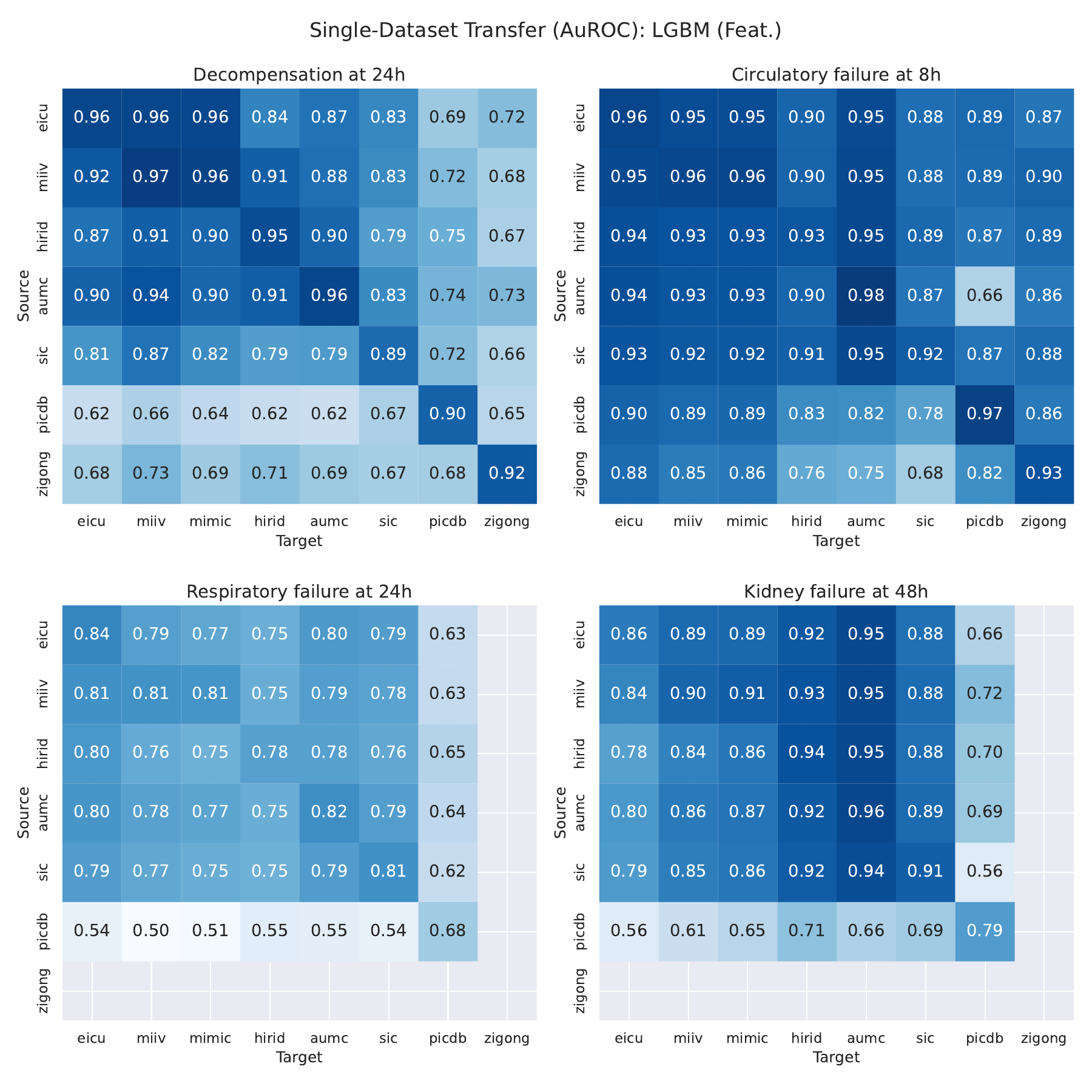}
        \caption{LGBM (Feat.) task-wise}
        \label{fig:single-center-transfer-lgbm}
    \end{subfigure}
\hfill
\begin{minipage}[t]{0.38\textwidth} \vspace{-8.5cm}
        \begin{subfigure}[t]{0.75\textwidth}
            \centering
            \includegraphics[width=\linewidth]{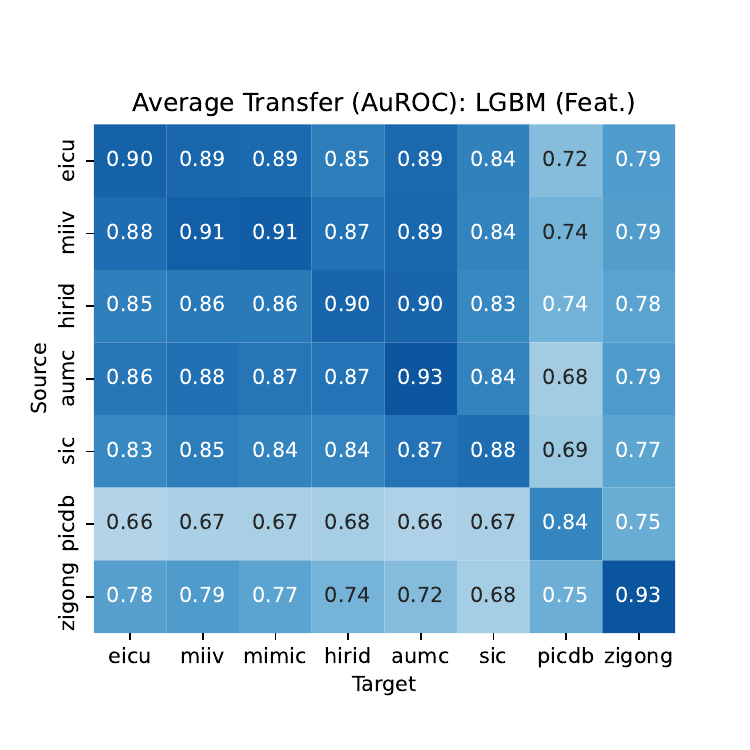}
            \caption{LGBM (Feat.) task-average}
            \label{fig:lgbm-avg-transfer}
        \end{subfigure}
\begin{subfigure}[b]{0.75\textwidth}
            \centering
            \includegraphics[width=\linewidth]{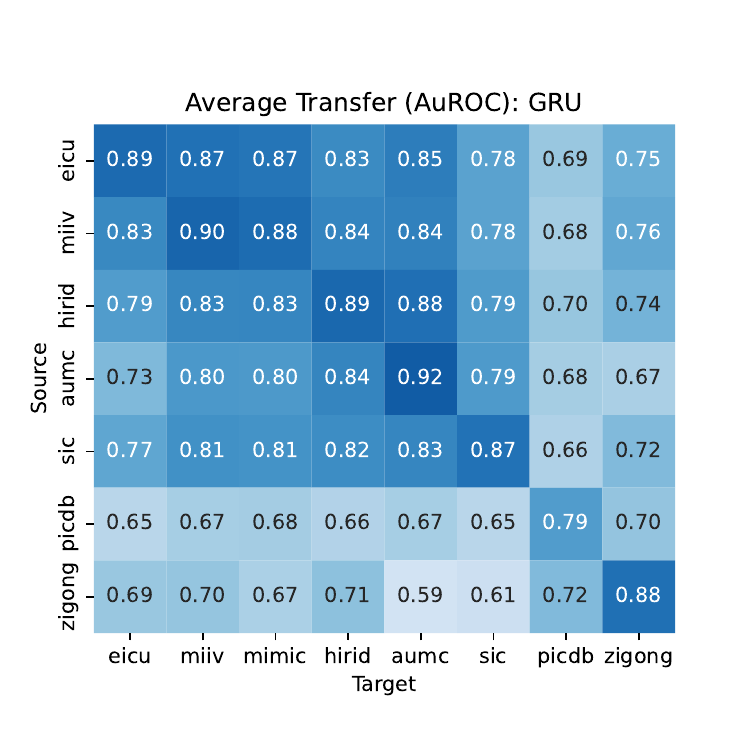} \caption{GRU task-average}
            \label{fig:gru-avg-transfer}
        \end{subfigure}
    \end{minipage}
    \caption{Single-center transfer performance heatmaps (AUROC). \cref{fig:single-center-transfer-lgbm} shows separate heatmaps for each task, while~\cref{fig:lgbm-avg-transfer,fig:gru-avg-transfer} show task-averaged performance. Further results are shown in~\cref{fig:transformer-mamba-avg-transfer} in \cref{sec:app_add_results}.}
    \label{fig:joint-heatmap}
\end{figure}

Here, in the single-center setting, we train a model on any single dataset and report test performance on the held-out dataset. In the multi-center hold-out setting, we train a model on all but the target dataset and then report test set performance on the held-out dataset. 

Results for out-of-distribution transfer for single- and multi-center training are presented in \cref{tab:benchmark-ood}. Hyperparameter optimization is always performed on the in-distribution validation sets corresponding to the collection of training sets.

For single-center experiments, we report the performance of the model and dataset that transferred the best. We do not consider this to be a realistic deployment scenario, but rather a reference point for the multi-center results. It verifies whether training on a single dataset that is similar to the target dataset is better than training on a collection of harmonized datasets. In real world, selecting such a training dataset would require either strong evidence of ``similarity'' (hard to justify as we observe transfer difficulties even within the same country) or considerable validation effort (technically infeasible for the majority of hospitals).

The multi-center hold-out is a more realistic setting as it does not require any prior knowledge or evaluation sets for the selection of the training set for transfer. It represents a deployment scenario, where a hospital with little data suitable for ML training uses and adapts the model built on open-access data.

\begin{table*}[tb]
\centering
\resizebox{\textwidth}{!}{\begin{tabular}{l l c c c | c c c c c c c}
    \toprule
    \bf Dataset & \bf Task & \multicolumn{3}{c}{\textbf{Single-Center}} & \multicolumn{7}{c}{\textbf{Multi-Center hold-out}} \\
    \cmidrule(lr){3-5} \cmidrule(lr){6-12} 
     & & AUROC & Best Model & Src. Data & 
     \rot{LR (Last Meas.)} & \rot{LGBM (Last Meas.)} & \rot{LGBM (Feat.)} & \rot{GRU} & \rot{Transformer} & \rot{Mamba}  & \rot{xLSTM} \\

\midrule
    \multirow{4}{*}{\norot{MIMIC-IV}} & Dec. 24h & \textbf{95.8} & LGBM (Feat.) & eICU & 91.5 & 94.0 & 95.6 & 94.1 & 94.0 & 93.8 & 94.5 \\
    & Circ. 8h  & \textbf{94.5} & LGBM (Feat.) & eICU & 91.6 & 92.9 & 94.5 & 93.6 & 93.5 & 93.3 & 93.5 \\
    & Resp. 24h & \textbf{78.5} & LGBM (Feat.) & eICU & 72.7 & 76.2 & 78.0 & 76.8 & 76.5 & 76.7 & 76.7 \\
    & Kidn. 48h & \textbf{89.1} & LGBM (Feat.) & eICU & 80.8 & 84.4 & 88.0 & 85.0 & 83.8 & 84.7 & 83.3 \\

\midrule
    \multirow{4}{*}{\norot{eICU}} & Dec. 24h & \textbf{92.3} & LGBM (Feat.) & MIMIC-IV & 86.3 & 90.9 & 92.2 & 90.3 & 90.5 & 90.5 & 90.4 \\
    & Circ. 8h  & 95.0 & LGBM (Feat.) & MIMIC-IV & 92.5 & 93.9 & \textbf{95.2} & 93.7 & 93.6 & 93.7 & 93.7 \\
    & Resp. 24h & \textbf{81.4} & LGBM (Feat.) & MIMIC-IV & 76.6 & 79.6 & 80.4 & 78.6 & 78.2 & 79.2 & 78.8 \\
    & Kidn. 48h & \textbf{83.7} & LGBM (Feat.) & MIMIC-IV & 71.9 & 78.7 & 82.2 & 77.5 & 77.4 & 77.9 & 77.2 \\

\midrule
    \multirow{4}{*}{\norot{HiRID}} & Dec. 24h & 91.1 & LGBM (Feat.) & UMCdb & 89.7 & 92.3 & \textbf{92.8} & 91.9 & 92.1 & 92.2 & 91.8 \\
    & Circ. 8h  & 90.7 & LGBM (Feat.) & SICdb    & 89.7 & 91.1 & \textbf{91.5} & 90.7 & 90.7 & 89.8 & 90.0 \\
    & Resp. 24h & 75.3 & LGBM (Feat.) & MIMIC-IV & 74.1 & 74.4 & 75.7 & \textbf{75.9} & 75.8 & 75.0 & 74.7 \\
    & Kidn. 48h & \textbf{93.2} & LGBM (Feat.) & MIMIC-IV & 89.9 & 92.3 & \textbf{93.2} & 92.3 & 92.0 & 91.1 & 90.7 \\

\midrule
    \multirow{4}{*}{\norot{UMCdb}} & Dec. 24h & 89.8 & LGBM (Feat.) & HiRID & 86.1 & 90.0 & \textbf{91.3} & 90.6 & 90.4 & 89.8 & 90.1 \\
    & Circ. 8h  & 96.4 & GRU & HiRID & 95.2 & 95.9 & 95.9 & 96.5 & 96.3 & 96.3 & \textbf{96.6} \\
    & Resp. 24h & \textbf{79.7} & LGBM (Feat.) & eICU & 77.0 & 78.1 & 78.1 & 79.2 & 77.7 & 76.4 & 76.7 \\
    & Kidn. 48h & 95.6 & LGBM (Last Meas.) & MIMIC-IV & 93.1 & 95.8 & \textbf{96.2} & 95.0 & 94.4 & 94.4 & 93.4 \\
    
\midrule
    \multirow{4}{*}{\norot{SICdb}} & Dec. 24h & 83.3 & LGBM (Feat.) & eICU & 79.7 & 84.3 & \textbf{84.4} & 83.4 & 83.0 & 82.3 & 81.6 \\
    & Circ. 8h  & 89.1 & LGBM (Feat.) & HiRID & 87.3 & 88.2 & 88.8 & 88.9 & \textbf{89.4} & 88.5 & 88.9 \\
    & Resp. 24h & 78.8 & LGBM (Feat.) & eICU & 76.5 & 78.2 & \textbf{79.0} & 77.7 & 77.2 & 77.6 & 77.5 \\
    & Kidn. 48h & 88.8 & LGBM (Feat.) & UMCdb & 84.2 & 87.8 & \textbf{90.0} & 85.7 & 85.9 & 85.5 & 85.5 \\

\midrule
    \multirow{4}{*}{\norot{PICdb}} & Dec. 24h & 75.0 & LGBM (Feat.) & HiRID & 67.9 & 74.3 & \textbf{79.6} & 66.1 & 66.0 & 62.6 & 61.5 \\
    & Circ. 8h  & \textbf{90.4} & GRU & MIMIC-IV & 88.3 & 89.3 & 87.5 & 87.6 & 87.1 & 86.4 & 86.6 \\
    & Resp. 24h & \textbf{70.8} & GRU & HiRID & 66.6 & 60.6 & 68.1 & 66.6 & 63.0 & 65.3 & 65.9 \\
    & Kidn. 48h & \textbf{71.9} & LGBM (Last Meas.) & HiRID & 57.6 & 67.7 & 68.5 & 63.0 & 61.2 & 56.7 & 69.0 \\

\midrule
    \multirow{2}{*}{\norot{Zigong}} & Dec. 24h & \textbf{72.8} & LGBM (Feat.) & UMCdb & 66.6 & 69.3 & 72.4 & 69.0 & 67.0 & 68.5 & 68.7 \\
    & Circ. 8h  & \textbf{91.4} & LGBM (Last Meas.) & MIMIC-IV & 89.0 & 88.6 & 89.0 & 87.1 & 86.5 & 85.2 & 84.7 \\

    \bottomrule
    \end{tabular}
}     
    \caption{Benchmarking results out-of-distribution (AUROC). Bold is best in each row (separately for single-cente and multi-center). Single-center results are an argmax over training datasets while testing on a hold-out dataset. Multi-center models are trained on all but the test dataset. All results show the mean over three different random initialization, except for LR models that are trained using convex optimization.}
\label{tab:benchmark-ood}
\end{table*} 
\subsection{Fine-tuning Study}

\begin{figure}[htbp]
    \centering
\begin{subfigure}[b]{0.45\textwidth}
        \centering
        \includegraphics[width=0.9\linewidth]{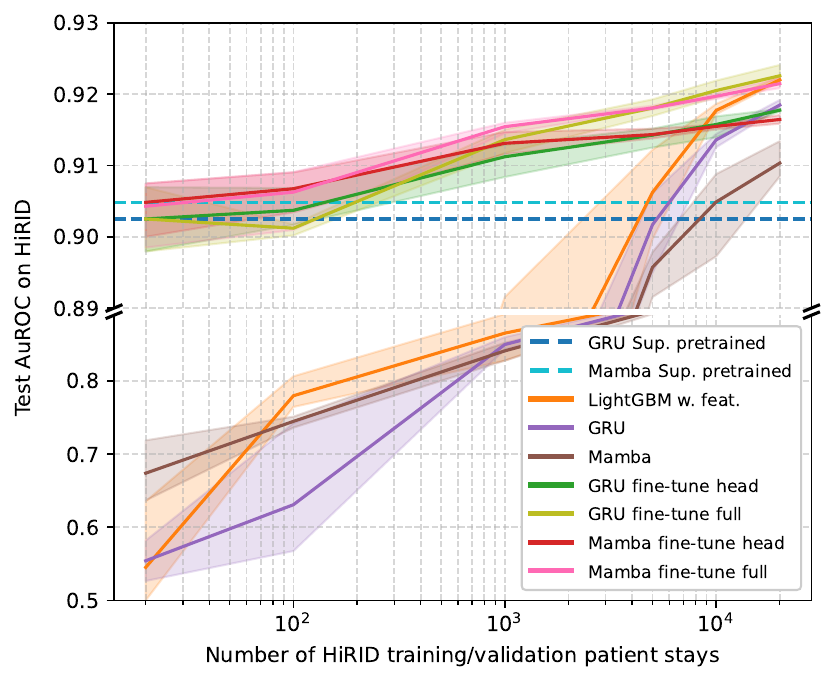}
        \caption{Circ. 8h (AUROC)}
        \label{fig:fine-tuning-hirid-auroc}
    \end{subfigure}
    \hfill
\begin{subfigure}[b]{0.45\textwidth}
        \centering
        \includegraphics[width=0.9\linewidth]{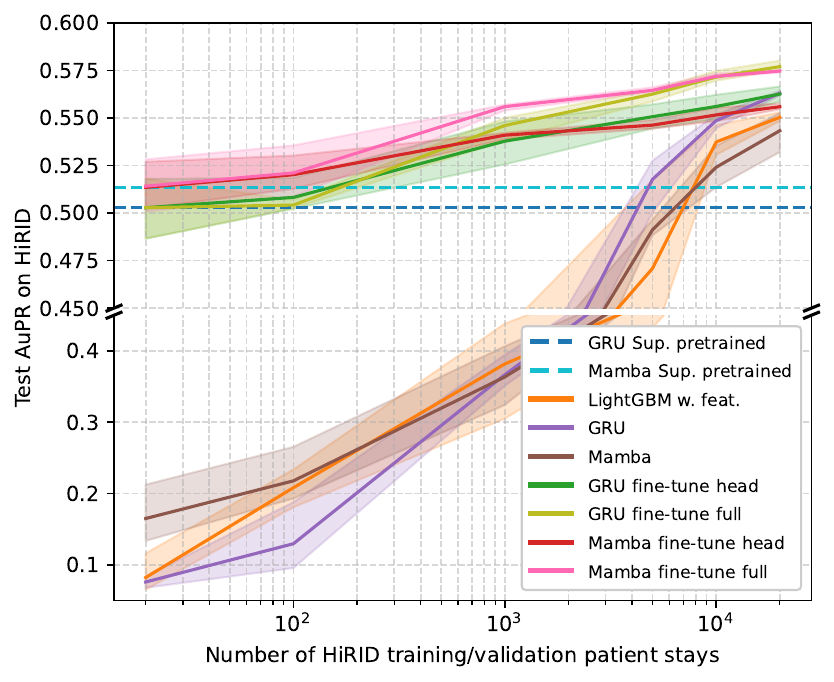}
        \caption{Circ. 8h (AUPRC)}
        \label{fig:fine-tuning-hirid-aupr}
    \end{subfigure}

    \begin{subfigure}[b]{0.45\textwidth}
        \centering
        \includegraphics[width=0.9\linewidth]{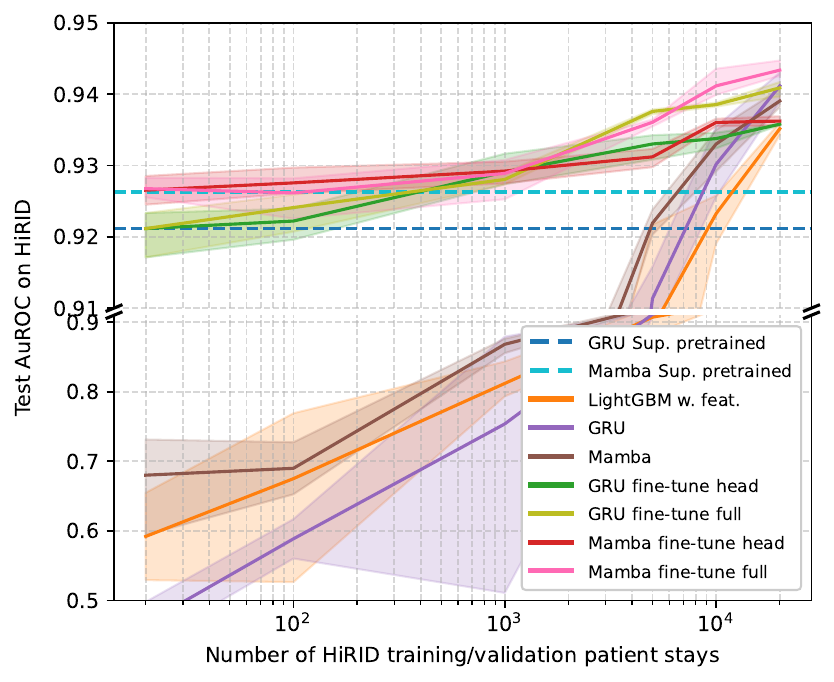}
        \caption{Dec. 24h (AUROC)}
        \label{fig:fine-tuning-hirid-auroc-decomp}
    \end{subfigure}
    \hfill
\begin{subfigure}[b]{0.45\textwidth}
        \centering
        \includegraphics[width=0.9\linewidth]{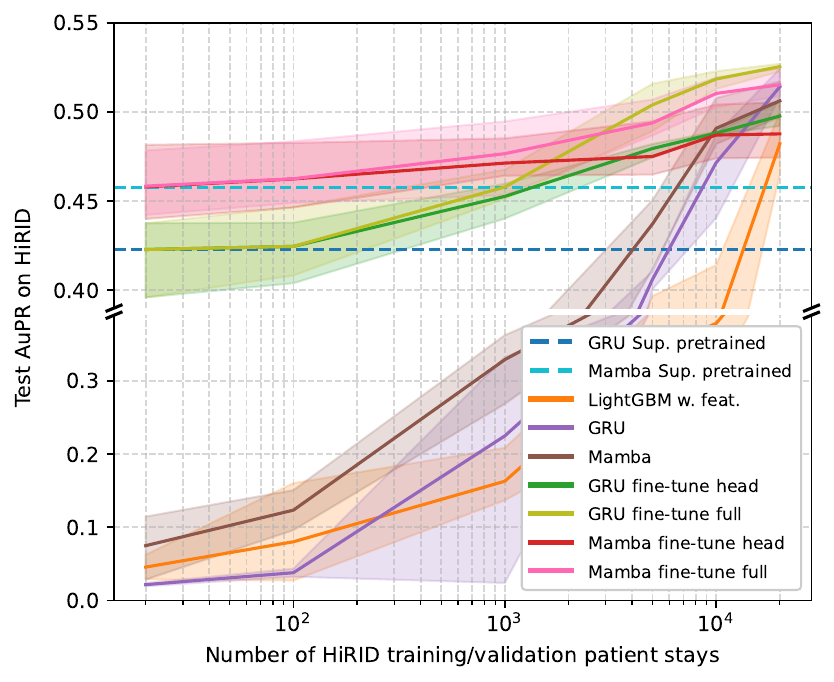}
        \caption{Dec. 24h (AUPRC)}
        \label{fig:fine-tuning-hirid-aupr-decomp}
    \end{subfigure}
    
    \caption{Supervised fine-tuning study performed on HiRID for circulatory failure prediction (\cref{fig:fine-tuning-hirid-auroc,fig:fine-tuning-hirid-aupr}) and decompensation (\cref{fig:fine-tuning-hirid-auroc-decomp,fig:fine-tuning-hirid-aupr-decomp}) by progressively increasing the number of admissions used for training or fine-tuning. \emph{GRU}, \emph{Mamba}, and \emph{LGBM w. feat.} are trained from scratch using HiRID data only. \emph{GRU/Mamba pretrained} is trained on all data excluding HiRID patients. \emph{GRU/Mamba fine-tuned (head/full)} is initialized with \emph{GRU/Mamba pretrained} and fine-tuned either across the full network or just the single linear logit head.}
    \label{fig:fine-tuning-study}
\end{figure}

We performed a first supervised pretraining and fine-tuning study, as shown in~\cref{fig:fine-tuning-study} (and~\cref{sec:app_add_results},~\cref{fig:fine-tuning-study-resp-kidney}), using the HiRID dataset~\citep{hirid}. We trained three models from scratch on a progressively increasing number of HiRID patients: LightGBM with extracted features as the best performing model for single-center results, GRU as the best performing deep sequence architecture, and Mamba as a more recent RNN architecture in comparison. Further, we pretrained a GRU/Mamba backbone on all other datasets in a supervised fashion and reported the zero-shot transfer performance without using any HiRID data. Finally, we initialized a GRU/Mamba network 
 with the aforementioned supervised pretrained weights and fine-tuned either the full network or only the linear logit head.

 \section{Discussion}

Overall, the transfer performance in- and out-of-distribution suggests that even without fine-tuning the resulting time series models are capable of performing early event prediction for relevant medical labels reasonably well.

In all settings, we see that gradient-boosted trees with feature extraction are the best-performing model across the board (see \cref{tab:benchmark-id,tab:benchmark-ood}). This is consistent with the previous findings~\citep{hyland2020early, yeche2021}.

At the same time, we note that the performance of deep models is often within just one or two AUROC points of the classical algorithms. For disposition prediction, the gap is even smaller, around a tenth of a point (\cref{tab:disposition}). In some cases, they manage to outperform tree-based alternatives, especially in multi-center settings. Different deep architectures are generally quite close to each other in terms of performance, without a clear leader emerging.

From the transfer heatmaps \cref{fig:joint-heatmap} we notice that (1) models generally transfer better in particular groups of datasets and (2) the transfer performance depends on the task.

For kidney failure, transfer works particularly well between HiRID, UMCdb, and SICdb (see bottom right \cref{fig:lgbm-avg-transfer}). We notice that these datasets are extracted at high resolution (five minutes, as opposed to an hour for other datasets, see \cref{tab:datasets_stats}). This suggests that recording resolution is an important factor for kidney failure prediction and its transferability.

On average, both LGBM and GRU generally transfer well between the eICU, MIMIC, HiRID, UMCdb, and SICdb datasets (\cref{fig:lgbm-avg-transfer,fig:gru-avg-transfer}). We hypothesize that this is due to their locality. These datasets originate from the USA and Europe, where clinical practices might be more similar than for example between the USA and China (PICdb and Zigong EHR datasets). This is reinforced by the particularly good transfer between eICU and MIMIC, both originating in the USA. 

The fine-tuning study (see~\cref{fig:fine-tuning-study} and~\cref{sec:app_add_results},~\cref{fig:fine-tuning-study-resp-kidney}) suggests that the models trained on the harmonized collection of datasets are able to generalize to a new, previously unseen dataset. They outperform a model trained from scratch for dataset sizes of up to tens of thousands of admissions. The practical implication is that training on publicly available datasets should be the go-to strategy for small and medium-scale studies on critical care time series. We also see that a fine-tuned GRU model performs better or is on par with the LGBM model trained from scratch on admission counts fewer and larger than 10,000. This result suggests that deep learning models might be a preferable choice when transferred from large datasets.
 \section{Conclusion}

In this work, we established the foundation for large-scale time series models on critical care data. We created the largest harmonized dataset that includes hospitals from three continents, incorporates treatment variables, and integrates data from both ICU and ED units. The dataset is further supported by a comprehensive transfer learning benchmark. 

Our results demonstrated that with access to an increased amount of carefully harmonized and labeled data, machine learning models are capable of generalizing when transferring across countries and continents, even without extensive fine-tuning. Notably, gradient-boosted trees with feature extraction consistently outperformed other models, although deep learning models came remarkably close, particularly in multi-center settings and for disposition prediction tasks. Importantly, our study highlights how dataset resolution and geographic origin influence transferability. Finally, fine-tuned models, trained on harmonized datasets, significantly improved performance on previously unseen data, especially for small and medium-sized datasets.

In future work, we plan to explore further improvements to the data coverage and training procedure to create the first foundation model for critical care time series and beyond.
 
\clearpage
\section*{Acknowledgments}

This work was supported as part of the Swiss AI Initiative (\href{https://swiss-ai.org}{\texttt{swiss-ai.org}}) by a grant from the Swiss National Supercomputing Centre (CSCS) under project ID \texttt{a02} on Alps.
Computational data analysis was performed at Leonhard Med (\url{https://sis.id.ethz.ch/services/sensitiveresearchdata/}) secure trusted research environment at ETH Zurich.

The project was supported by grant \#2022-278 of the Strategic Focus Area ``Personalized Health and Related Technologies (PHRT)'' of the ETH Domain (Swiss Federal Institutes of Technology) and ETH Core funding (to G.R.).

Malte Londschien was supported by the ETH Foundations of Data Science and the ETH AI Center. Fedor Sergeev was supported by grant \#902 of the Strategic Focus Area ``Personalized Health and Related Technologies (PHRT)'' of the ETH Domain (Swiss Federal Institutes of Technology). Fedor Sergeev thanks Vincent Fortuin for his helpful suggestions to improve the manuscript. Daphné Chopard received funding from grant \#2021-911 of the Strategic Focal Area ``Personalized Health and Related Technologies (PHRT)'' of the ETH Domain (Swiss Federal Institutes of Technology).

\bibliography{icufm}
\bibliographystyle{plainnat}

\clearpage
\appendix
\section{Data}
\label{sec:app_data}

\subsection{Datasets}

The overview of the datasets is shown in \cref{tab:datasets_stats}. We do not harmonize RICD~\citep{vorontsov2024virchowmillionslidedigitalpathology} as it is not free access (a separate contract and payment are required). Other datasets are available via PhysioNet \citep{goldberger2000physiobank} or directly from the providers.

\begin{table}[htb]
\centering
\resizebox{\textwidth}{!}{\begin{tabular}{clrrrrrrrrrrr}
    \toprule
     &  & \multicolumn{1}{c}{\fullrot{MIMIC-III}} & \multicolumn{1}{c}{\fullrot{MIMIC-IV}} & \multicolumn{1}{c}{\fullrot{eICU}} & \multicolumn{1}{c}{\fullrot{UMCdb\textsuperscript{\S}}} & \multicolumn{1}{c}{\fullrot{HiRID}} & \multicolumn{1}{c}{\fullrot{SICdb}} & \multicolumn{1}{c}{\fullrot{PICdb}} & \multicolumn{1}{c}{\fullrot{Zigong EHR}} & \multicolumn{1}{c}{\fullrot{MIMIC-IV-ED}}  & \multicolumn{1}{c}{\fullrot{\sout{RICD}}} \\ \midrule
     \multirow{9}{*}{\fullrot{General\textsuperscript{\dag}}} & Time & 2001-12 & 2008-19 & 2014-15 & 2003-16 & 2008-16 & 2013-21 & 2010-18 & 2019-20 & 2011-19 & 2017-23 \\
     & Country & US & US & US & NL & CH & AU & CN & CN & US & RU \\
     & Easy access\textsuperscript{*} & $+$ & $+$ & $+$ & $\pm$ & $\pm$ & $\pm$ & $+$ & $+$ & $+$  & $-$ \\
     & Max resolution, min & 60 & 60 & 60 & 60 & 2 & 1 & 5 & 60 & -- & 6 \\
     & Admissions & 61532 & 76540 & 200859 & 23106 & 33905 & 27386 & 13499 & 2790 & $\sim$ 425000 & 3291 \\
     & Patients & 46476 & 53150 & -- & 20109 & -- & -- & 12881 & 2790 & --  & 2562 \\
     & Mean LoS, days & 2.1 & 11 & 1.57 & 1.08 & 0.95 & 3.5 & 9.3 & 4 & -- & 32 \\
     & Mean age, years & 65.8 & 64.7 & 65 & 65 & 65 & -- & 0.8 & 69.2 & -- & 57.8 \\
    & Mortality, \% & 8.5 & 11.6 & 9.94 & 12.05 & 6.52 & 3.45 & 6.9 & 5.77 & -- & 12.31\\\addlinespace
     \multirow{-2}{*}{\fullrot{Extracted}} & Resolution, min & 60 & 60 & 60 & 5 & 5 & 5 & 60 & 60 & 60 & -- \\
     & Admissions & 53713 & 70831 & 183695 & 22889 & 33558 & 24522 & 13295 & 2525 & 177714 & -- \\ \addlinespace
     \multirow{5}{*}{\fullrot{Label, \%}} & Decompensation 24h          &  8 &  7 &  5 & 10 &  6 &  5 &  7 & 43 & -- & -- \\
     & Circulatory 8h  & 14 & 19 &  7 & 23 & 31 & 30 & 16 & 32 & -- & -- \\
     & Respiratory 24h & 22 & 25 & 14 & 45 & 45 & 54 &  2 &  0 & -- & -- \\
     & Kidney 48h      &  5 &  7 &  7 &  5 &  3 &  5 &  1 &  0 & -- & -- \\
     & ED Disposition  & -- & -- & -- & -- & -- & -- & -- & -- & 51 & -- \\ \bottomrule
\end{tabular}}
\caption{Datasets overview.\smallskip\\\footnotesize \textsuperscript{*} Denoted as $+$ if only a CITI certificate\protect\footnotemark and $\pm$ if additional provider approval is required, $-$ otherwise.\\ \textsuperscript{\S} Shortened AmsterdamUMCdb to ``UMCdb''.\\ \textsuperscript{\dag} We consider ED admissions for MIMIC-IV-ED, and ICU admissions for other datasets.}
\label{tab:datasets_stats}
\end{table}

\footnotetext{Certificate ``Data or Specimens Only Research'' from the Collaborative Institutional Training Initiative (CITI) program: \url{physionet.org/about/citi-course/}}

\subsection{Inclusion criteria}
\label{sec:inclusion-criteria}

We consider patient stays that after extraction have 
\begin{itemize}
    \item a valid admission and discharge time,
    \item a valid length of stay (LoS) that is longer than 4 fours,
    \item a maximum gap between measurements smaller than 48 hours,
    \item and more than 4 measurements.
\end{itemize}

Compared to \citet{van2023yet}, we broaden the inclusion criteria by reducing the LoS requirement from 6 to 4 hours and increasing the allowed maximum gap between measurements from 12 to 48 hours.

By including as many patients as possible, we aim to create a more general version of the dataset, that can be further trimmed down for specific studies. Additionally, a wide range of stays can improve the generalizability of predictive models.

\subsection{Concepts for treatment variables}

In \cref{tab:metavariables_literature} we present concepts for treatments that were identified as important in the ML literature with their clinical importance. The choice of concepts balances granularity, missingness, and time effort, to incorporate as much of the signal from the data as possible while keeping missingness across datasets low and the variable labeling feasible for the medical experts.

\begin{table}[htb]
\centering
\begin{adjustbox}{width=1.0\textwidth}
    \renewcommand{\arraystretch}{1.2}
    \begin{tabular}{@{} llll lc cccc cccc @{}}
    \toprule
        \multicolumn{4}{c}{\textbf{Variables}} & \multicolumn{2}{c}{\textbf{Clinical importance}} & \multicolumn{4}{c}{\textbf{Task importance\textsuperscript{*}}} & \multicolumn{4}{c}{\textbf{Literature Importance\textsuperscript{\dag}}} \\
        \cmidrule(r){1-4} \cmidrule(lr){5-6} \cmidrule(lr){7-10} \cmidrule(l){11-14}
        \multicolumn{1}{c}{\textbf{Meta variable}} & \multicolumn{1}{c}{\textbf{Type}} & \multicolumn{1}{c}{\textbf{Organ system}} & \multicolumn{1}{c}{\textbf{Group}} & \multicolumn{1}{c}{\textbf{Included as}} & \textbf{Priority} & \textbf{Circ.} & \textbf{Resp.} & \textbf{Kidn.} & \textbf{Sepsis} & \textbf{CircEWS} & \textbf{RMS} & \textbf{KDIGO} & \textbf{Moor} \\ 
    \midrule
        Dobutamine & {\color[HTML]{073763} Drug} & Cardiovascular & Vasopressor / Inotropes & {\color[HTML]{741B47} Rate \& ind.} & {\color[HTML]{CC0000} High} & \cmark & \cmark & \cmark & \cmark & \cmark & \xmark & \xmark & \cmark \\
        Levosimendan & {\color[HTML]{073763} Drug} & Cardiovascular & Vasopressor / Inotropes & {\color[HTML]{741B47} Rate \& ind.} & {\color[HTML]{CC0000} High} & \cmark & \xmark & \cmark & \xmark & \cmark & \xmark & \cmark & \xmark \\
        Norepinephrine & {\color[HTML]{073763} Drug} & Cardiovascular & Vasopressor / Inotropes & {\color[HTML]{741B47} Rate \& ind.} & {\color[HTML]{CC0000} High} & \xmark & \cmark & \xmark & \cmark & \xmark & \cmark & \xmark & \cmark \\
        Epinephrine & {\color[HTML]{073763} Drug} & Cardiovascular & Vasopressor / Inotropes & {\color[HTML]{741B47} Rate \& ind.} & {\color[HTML]{CC0000} High} & \xmark & \cmark & \xmark & \cmark & \xmark & \cmark & \xmark & \cmark \\
        Milrinone & {\color[HTML]{073763} Drug} & Cardiovascular & Vasopressor / Inotropes & {\color[HTML]{741B47} Rate \& ind.} & {\color[HTML]{CC0000} High} & \cmark & \xmark & \xmark & \xmark & \cmark & \xmark & \xmark & \xmark \\
        Theophylline & {\color[HTML]{073763} Drug} & Cardiovascular & Vasopressor / Inotropes & {\color[HTML]{741B47} Rate \& ind.} & {\color[HTML]{CC0000} High} & \cmark & \xmark & \xmark & \xmark & \cmark & \xmark & \xmark & \xmark \\
        Dopamine & {\color[HTML]{073763} Drug} & Cardiovascular & Vasopressor / Inotropes & {\color[HTML]{741B47} Rate \& ind.} & {\color[HTML]{CC0000} High} & \xmark & \xmark & \xmark & \cmark & \xmark & \xmark & \xmark & \cmark \\
        Vasopressin & {\color[HTML]{073763} Drug} & Cardiovascular & Vasopressor / Inotropes & {\color[HTML]{741B47} Rate \& ind.} & {\color[HTML]{CC0000} High} & \xmark & \xmark & \xmark & \xmark & \xmark & \xmark & \xmark & \xmark \\
        Heparin & {\color[HTML]{073763} Drug} & Cardiovascular & Anticoagulants & {\color[HTML]{741B47} Rate \& ind.} & {\color[HTML]{CC0000} High} & \xmark & \cmark & \cmark & \xmark & \xmark & \cmark & \cmark & \xmark \\
        Propofol & {\color[HTML]{073763} Drug} & Nervous & Sedatives / Anxiolytics & {\color[HTML]{741B47} Rate \& ind.} & {\color[HTML]{CC0000} High} & \xmark & \cmark & \xmark & \xmark & \xmark & \cmark & \xmark & \xmark \\
        Benzodiacepine & {\color[HTML]{073763} Drug} & Nervous & Sedatives / Anxiolytics & {\color[HTML]{741B47} Rate \& ind.} & {\color[HTML]{CC0000} High} & \xmark & \cmark & \xmark & \xmark & \xmark & \cmark & \xmark & \xmark \\
        Loop diuretic & {\color[HTML]{073763} Drug} & Renal & Diuretics & {\color[HTML]{741B47} Rate \& ind.} & {\color[HTML]{CC0000} High} & \xmark & \cmark & \cmark & \xmark & \xmark & \cmark & \cmark & \xmark \\
        Other sedatives & {\color[HTML]{073763} Drug} & Nervous & Sedatives / Anxiolytics & {\color[HTML]{0B5394} Indicator} & {\color[HTML]{CC0000} High} & \xmark & \xmark & \xmark & \xmark & \xmark & \xmark & \xmark & \xmark \\
        Opiate painkillers & {\color[HTML]{073763} Drug} & Nervous & Pain killers & {\color[HTML]{0B5394} Indicator} & {\color[HTML]{CC0000} High} & \xmark & \xmark & \cmark & \xmark & \xmark & \xmark & \cmark & \xmark \\
        Non-opioid analgesic & {\color[HTML]{073763} Drug} & Nervous & Pain killers & {\color[HTML]{0B5394} Indicator} & {\color[HTML]{CC0000} High} & \cmark & \xmark & \xmark & \xmark & \cmark & \xmark & \xmark & \xmark \\
        Paralytics & {\color[HTML]{073763} Drug} & Nervous & Paralyzing & {\color[HTML]{0B5394} Indicator} & {\color[HTML]{CC0000} High} & \xmark & \xmark & \xmark & \xmark & \xmark & \xmark & \xmark & \xmark \\
        Administration of antibotics & {\color[HTML]{073763} Drug} & Infectious & Antibiotics & {\color[HTML]{0B5394} Indicator} & {\color[HTML]{CC0000} High} & \xmark & \xmark & \cmark & \xmark & \xmark & \xmark & \cmark & \xmark \\
        Insulin & {\color[HTML]{073763} Drug} & Endocrine & Insulin & {\color[HTML]{0B5394} Indicator} & {\color[HTML]{CC0000} High} & \xmark & \cmark & \xmark & \xmark & \xmark & \cmark & \xmark & \xmark \\
        Anti delirant medi & {\color[HTML]{073763} Drug} & Nervous & Anti delirant medi & {\color[HTML]{0B5394} Indicator} & {\color[HTML]{B45F06} Med} & \xmark & \xmark & \cmark & \xmark & \xmark & \xmark & \cmark & \xmark \\
        Other diuretics & {\color[HTML]{073763} Drug} & Renal & Diuretics & {\color[HTML]{0B5394} Indicator} & {\color[HTML]{B45F06} Med} & \xmark & \xmark & \xmark & \xmark & \xmark & \xmark & \xmark & \xmark \\
        Other anticoagulants & {\color[HTML]{073763} Drug} & Cardiovascular & Anticoagulants & {\color[HTML]{0B5394} Indicator} & {\color[HTML]{B45F06} Med} & \xmark & \xmark & \xmark & \xmark & \xmark & \xmark & \xmark & \xmark \\
        Vasodilators & {\color[HTML]{073763} Drug} & Cardiovascular & Antihypertensive + Vasodilators & {\color[HTML]{0B5394} Indicator} & {\color[HTML]{B45F06} Med} & \xmark & \xmark & \xmark & \xmark & \xmark & \xmark & \xmark & \xmark \\
        Antiarrhythmics & {\color[HTML]{073763} Drug} & Cardiovascular & Antiarrhythmic & {\color[HTML]{0B5394} Indicator} & {\color[HTML]{B45F06} Med} & \xmark & \xmark & \xmark & \xmark & \xmark & \xmark & \xmark & \xmark \\
        Packed red blood cells & {\color[HTML]{741B47} Blood} & Cardiovascular / Renal & Infusion of blood products & {\color[HTML]{0B5394} Indicator} & {\color[HTML]{B45F06} Med} & \xmark & \xmark & \cmark & \xmark & \xmark & \xmark & \cmark & \xmark \\
        FFp & {\color[HTML]{741B47} Blood} & Cardiovascular / Renal & Infusion of blood products & {\color[HTML]{0B5394} Indicator} & {\color[HTML]{B45F06} Med} & \xmark & \xmark & \xmark & \xmark & \xmark & \xmark & \xmark & \xmark \\
        Platelets & {\color[HTML]{741B47} Blood} & Cardiovascular / Renal & Infusion of blood products & {\color[HTML]{0B5394} Indicator} & {\color[HTML]{B45F06} Med} & \xmark & \xmark & \xmark & \xmark & \xmark & \xmark & \xmark & \xmark \\
        Albumin & {\color[HTML]{741B47} Blood} & Cardiovascular / Renal & Infusion of blood products & {\color[HTML]{0B5394} Indicator} & {\color[HTML]{B45F06} Med} & \xmark & \xmark & \xmark & \xmark & \xmark & \xmark & \xmark & \xmark \\
        Fluid administration & {\color[HTML]{351C75} Feeding / Electrolyte} & Gastrointestinal / Renal & Electrolytes & {\color[HTML]{0B5394} Indicator} & {\color[HTML]{B45F06} Med} & \xmark & \xmark & \cmark & \xmark & \xmark & \cmark & \xmark & \xmark \\
        Electrolytes-Phosphate & {\color[HTML]{351C75} Feeding / Electrolyte} & Gastrointestinal / Renal & Electrolytes & None & {\color[HTML]{38761D} Low} & \xmark & \xmark & \xmark & \xmark & \xmark & \xmark & \xmark & \xmark \\
        Electrolytes-Kalium & {\color[HTML]{351C75} Feeding / Electrolyte} & Gastrointestinal / Renal & Electrolytes & None & {\color[HTML]{38761D} Low} & \xmark & \xmark & \cmark & \xmark & \xmark & \xmark & \xmark & \xmark \\
        Elektrolytes-Mg & {\color[HTML]{351C75} Feeding / Electrolyte} & Gastrointestinal / Renal & Electrolytes & None & {\color[HTML]{38761D} Low} & \xmark & \xmark & \cmark & \xmark & \xmark & \xmark & \cmark & \xmark \\
        Enteral feeding & {\color[HTML]{351C75} Feeding / Electrolyte} & Gastrointestinal / Renal & Feeding & None & {\color[HTML]{38761D} Low} & \xmark & \xmark & \cmark & \xmark & \xmark & \xmark & \cmark & \xmark \\
        Parenteral feeding & {\color[HTML]{351C75} Feeding / Electrolyte} & Gastrointestinal / Renal & Feeding & None & {\color[HTML]{38761D} Low} & \xmark & \xmark & \cmark & \xmark & \xmark & \xmark & \cmark & \xmark \\
        Glucose & {\color[HTML]{073763} Drug} & Endocrine & Glucose & None & {\color[HTML]{38761D} Low} & \xmark & \xmark & \xmark & \xmark & \xmark & \xmark & \xmark & \xmark \\
        Antiepileptic & {\color[HTML]{073763} Drug} & Nervous & Antiepileptic & None & {\color[HTML]{38761D} Low} & \xmark & \xmark & \xmark & \xmark & \xmark & \xmark & \xmark & \xmark \\
        Inhalation & {\color[HTML]{073763} Drug} & Respiratory & Inhalation & None & {\color[HTML]{38761D} Low} & \xmark & \xmark & \xmark & \xmark & \xmark & \xmark & \xmark & \xmark \\
        Platelet inhibitors & {\color[HTML]{073763} Drug} & Cardiovascular & Platelet inhibitors & None & {\color[HTML]{38761D} Low} & \xmark & \xmark & \cmark & \xmark & \xmark & \xmark & \cmark & \xmark \\
        Desmopressin & {\color[HTML]{073763} Drug} & Cardiovascular & Vasopressor / Inotropes & None & {\color[HTML]{38761D} Low} & \xmark & \xmark & \xmark & \xmark & \xmark & \xmark & \xmark & \xmark \\
        Inhalation & {\color[HTML]{073763} Drug} & Respiratory & Inhalation & None & {\color[HTML]{38761D} Low} & \xmark & \xmark & \xmark & \xmark & \xmark & \xmark & \xmark & \xmark \\
        Immunmodulation & {\color[HTML]{073763} Drug} & Immune & Immunmodulation & None & {\color[HTML]{38761D} Low} & \xmark & \xmark & \xmark & \xmark & \xmark & \xmark & \xmark & \xmark \\
        Laxatives & {\color[HTML]{073763} Drug} & Gastrointestinal & Laxatives & None & {\color[HTML]{38761D} Low} & \xmark & \xmark & \cmark & \xmark & \xmark & \xmark & \cmark & \xmark \\
        Peritoneal dialysis & {\color[HTML]{741B47} Blood} & Cardiovascular / Renal & Dialysis & None & {\color[HTML]{38761D} Low} & \xmark & \xmark & \cmark & \xmark & \xmark & \xmark & \cmark & \xmark \\
        Infusion of blood products & {\color[HTML]{741B47} Blood} & Cardiovascular / Renal & Blood products & None & {\color[HTML]{38761D} Low} & \xmark & \xmark & \cmark & \xmark & \xmark & \xmark & \xmark & \xmark \\
        Supplemental oxygen & {\color[HTML]{134F5C} Ventilator} & Respiratory & Respirator settings & {\color[HTML]{741B47} Rate} & {\color[HTML]{B45F06} Med} & \cmark & \cmark & \xmark & \xmark & \cmark & \cmark & \xmark & \xmark \\
        Other ventilator settings & {\color[HTML]{134F5C} Ventilator} & Respiratory & Respirator settings & None & {\color[HTML]{38761D} Low} & \xmark & \cmark & \cmark & \cmark & \xmark & \cmark & \xmark & \cmark \\
    \bottomrule
    \end{tabular}\end{adjustbox}
\caption{Treatment concepts and their importance in clinical practice and ML literature.\smallskip\\ \textsuperscript{*} \footnotesize Circ.\ is circulatory, Resp.\ is respiratory, and Kidn.\ is kidney failure.\\ \textsuperscript{\dag} CircEWS \citep{hyland2020early}, RMS \citep{huser2024medrxiv_respiratory}, KDIGO \citep{lyu2024empirical}, Moor \citep{moor2023predicting}.\normalsize}
\label{tab:metavariables_literature}
\end{table} 
The statistics for the new concepts covering medications are shown in \cref{tab:metavariables_stats}. We note that we include all medications as indicators, and, for the most important ones, rates if possible to compute (e.g., the information on dosage is included with a convertible unit and appropriate time information is available). 

We labeled but did not include the data in the MIMIC-III prescriptions table because it only specifies the prescription and not the drug administration.

\begin{table}[htb]
\centering
\resizebox{\textwidth}{!}{\begin{tabular}{@{}cllrrrrrrrrrr@{}}
    \toprule
     & \multicolumn{1}{c}{Concept name} & \multicolumn{1}{c}{Abbreviation} & \rot{MIMIC-IV} & \rot{HiRID} & \rot{MIMIC-III} & \rot{PICdb} & \rot{SICdb} & \rot{Zigong EHR} & \rot{eICU} & \rot{UMCdb} & \rot{MIMIC-IV-ED} & Total \\ \midrule
    \multirow{12}{*}{\rotatebox[origin=c]{90}{Rates \& indicators}} & dobutamine & dobu & 1 & 1 & 4 & 1 & 3 & 1 & 13 & 1 & 2 & 27 \\
     & levosimendan & levo & 0 & 1 & 0 & 0 & 1 & 1 & 0 & 0 & 0 & 3 \\
     & norepinephrine & norepi & 1 & 5 & 3 & 1 & 5 & 1 & 41 & 1 & 6 & 64 \\
     & epinephrine & epi & 2 & 5 & 8 & 1 & 1 & 1 & 20 & 1 & 10 & 49 \\
     & milrinone & milirin & 1 & 2 & 2 & 2 & 1 & 1 & 10 & 0 & 2 & 21 \\
     & theophylline & teophyllin & 1 & 6 & 3 & 2 & 1 & 2 & 6 & 4 & 0 & 25 \\
     & dopamine & dopa & 1 & 0 & 7 & 1 & 1 & 1 & 17 & 1 & 2 & 31 \\
     & vasopressin & adh & 1 & 2 & 17 & 0 & 1 & 0 & 22 & 0 & 3 & 46 \\
     & heparin & hep & 4 & 3 & 52 & 1 & 1 & 2 & 75 & 2 & 8 & 148 \\
     & propofol & prop & 2 & 13 & 4 & 1 & 5 & 3 & 39 & 1 & 3 & 71 \\
     & benzodiacepine & benzdia & 3 & 18 & 19 & 7 & 9 & 7 & 110 & 13 & 45 & 231 \\
     & loop diuretic & loop\_diur & 3 & 8 & 9 & 2 & 4 & 4 & 62 & 2 & 7 & 101 \\\addlinespace
    \multirow{16}{*}{\rotatebox[origin=c]{90}{Indicators}} & other sedatives & sed & 9 & 5 & 36 & 3 & 16 & 7 & 57 & 8 & 24 & 165 \\
     & opiate painkiller & op\_pain & 7 & 32 & 52 & 9 & 21 & 12 & 309 & 13 & 86 & 541 \\
     & non-opioid analgesic & nonop\_pain & 2 & 32 & 9 & 14 & 12 & 14 & 163 & 11 & 38 & 295 \\
     & paralytic & paral & 7 & 0 & 56 & 2 & 3 & 4 & 100 & 5 & 7 & 184 \\
     & antibotics & abx & 55 & 125 & 208 & 164 & 49 & 90 & 330 & 53 & 99 & 1173 \\
     & insulin & ins & 8 & 5 & 20 & 6 & 7 & 13 & 132 & 7 & 6 & 204 \\
     & fluid administration & fluid & 7 & 2 & 59 & 0 & 23 & 14 & 297 & 9 & 7 & 418 \\
     & packed red blood cells & inf\_rbc & 3 & 2 & 3 & 0 & 0 & 0 & 14 & 1 & 0 & 23 \\
     & fresh frozen plasma & ffp & 3 & 2 & 8 & 0 & 0 & 0 & 7 & 1 & 0 & 21 \\
     & platelets & plat & 3 & 2 & 3 & 0 & 1 & 0 & 7 & 1 & 0 & 17 \\
     & albumin infusion & inf\_alb & 2 & 0 & 14 & 1 & 4 & 2 & 57 & 1 & 0 & 81 \\
     & anti deliriant & anti\_delir & 1 & 16 & 0 & 1 & 0 & 2 & 9 & 1 & 4 & 34 \\
     & other diuretics & oth\_diur & 1 & 10 & 28 & 3 & 13 & 1 & 46 & 6 & 10 & 118 \\
     & other anticoagulants & anti\_coag & 8 & 18 & 43 & 6 & 15 & 16 & 161 & 14 & 20 & 301 \\
     & antihypertensive and vasodilators & vasod & 11 & 78 & 62 & 14 & 65 & 53 & 313 & 41 & 84 & 721 \\
     & antiarrhythmic & anti\_arrhythm & 14 & 8 & 23 & 8 & 14 & 8 & 95 & 11 & 27 & 208 \\\addlinespace
    \midrule[\lightrulewidth]
    \multicolumn{1}{l}{} & Used &  & 442 & 891 & 9458 & 719 & 1595 & 1077 & 9763 & 1113 & 1085 & 26143 \\
    \multicolumn{1}{l}{} & Not used &  & 292 & 492 & 8751 & 470 & 1332 & 818 & 7278 & 908 & 602 & 20943 \\
    \multicolumn{1}{l}{} & Total &  & 453 & 893 & 9503 & 720 & 1608 & 1078 & 9790 & 1117 & 1102 & 26264 \\ \bottomrule
\end{tabular}}
\caption{Presence of medication concepts across datasets}
\label{tab:metavariables_stats}
\end{table} 
\clearpage
\subsection{Concept reference table}

The full concept (or variable) reference table is shown in \cref{tab:variables}. It includes 141 variables: 6 static demographic, 80 observations, and 55 treatment variables. 

\smaller[2]
\begin{longtable}[c]{@{}lllllll@{}}
\toprule
\multicolumn{1}{c}{Tag} & \multicolumn{1}{c}{Name} & \multicolumn{1}{c}{Type} & \multicolumn{1}{c}{Organ System} & \multicolumn{1}{c}{Unit} \\* \midrule
\endfirsthead
\endhead
\bottomrule
\endfoot
\endlastfoot
map & Mean Arterial Blood Pressure & observation & circulatory & mmHg \\
lact & Lactate & observation & circulatory & mmol/L \\
age & Age & demographic & None & years \\
weight & Weight & demographic & None & kg \\
sex & Sex & demographic & None & categorical \\
height & Height & demographic & None & cm \\
hr & Heart Rate & observation & circulatory & bpm \\
fio2 & FiO2 & observation & respiratory & \% \\
resp & Respiratory Rate & observation & respiratory & insp/min \\
temp & Temperature & observation & infection & C \\
crea & Creatinine & observation & metabolic\_renal & mg/dL \\
urine\_rate & Urine Rate Per Hour & observation & metabolic\_renal & mL/h \\
po2 & Partial Pressure Of Oxygen & observation & respiratory & mmHg \\
ethnic & Ethnic Group & demographic & None & categorical \\
alb & Albumin & observation & gastrointestinal & g/dL \\
alp & Alkaline Phosphatase & observation & gastrointestinal & IU/L \\
alt & Alanine Aminotransferase & observation & gastrointestinal & IU/L \\
ast & Aspartate Aminotransferase & observation & gastrointestinal & IU/L \\
be & Base Excess & observation & metabolic\_renal & mmol/l \\
bicar & Bicarbonate & observation & metabolic\_renal & mmol/l \\
bili & Total Bilirubin & observation & gastrointestinal & mg/dL \\
bili\_dir & Bilirubin Direct & observation & gastrointestinal & mg/dL \\
bnd & Band Form Neutrophils & observation & infection & \% \\
bun & Blood Urea Nitrogen & observation & metabolic\_renal & mg/dL \\
ca & Calcium & observation & metabolic\_renal & mg/dL \\
cai & Calcium Ionized & observation & metabolic\_renal & mmol/L \\
ck & Creatine Kinase & observation & circulatory & IU/L \\
ckmb & Creatine Kinase MB & observation & circulatory & ng/mL \\
cl & Chloride & observation & metabolic\_renal & mmol/l \\
crp & C-Reactive Protein & observation & infection & mg/L \\
dbp & Diastolic Blood Pressure & observation & circulatory & mmHg \\
fgn & Fibrinogen & observation & circulatory & mg/dL \\
glu & Glucose & observation & metabolic\_renal & mg/dL \\
hgb & Hemoglobin & observation & circulatory & g/dL \\
inr\_pt & Prothrombin & observation & circulatory & INR \\
k & Potassium & observation & metabolic\_renal & mmol/l \\
lymph & Lymphocytes & observation & infection & \% \\
methb & Methemoglobin & observation & circulatory & \% \\
mg & Magnesium & observation & metabolic\_renal & mg/dL \\
na & Sodium & observation & metabolic\_renal & mmol/l \\
neut & Neutrophils & observation & infection & \% \\
pco2 & CO2 Partial Pressure & observation & respiratory & mmHg \\
ph & pH Of Blood & observation & metabolic\_renal & pH \\
phos & Phosphate & observation & metabolic\_renal & mg/dL \\
plt & Platelet Count & observation & circulatory & G/l \\
ptt & Partial Thromboplastin Time & observation & circulatory & sec \\
sbp & Systolic Blood Pressure & observation & circulatory & mmHg \\
tnt & Troponin T & observation & circulatory & ng/mL \\
wbc & White Blood Cell Count & observation & infection & G/l \\
basos & Basophils & observation & infection & \% \\
eos & Eosinophils & observation & infection & \% \\
mgcs & Glasgow Comma Scale Motor & observation & neuro & categorical \\
tgcs & Glasgow Comma Scale Total & observation & neuro & categorical \\
vgcs & Glasgow Comma Scale Verbal & observation & neuro & categorical \\
egcs & Glasgow Comma Scale Eye & observation & neuro & categorical \\
hct & Hematocrit & observation & circulatory & \% \\
rbc & Red Blood Cell Count & observation & circulatory & m/uL \\
tri & Troponin I & observation & circulatory & ng/mL \\
etco2 & Endtital CO2 & observation & respiratory & mmHg \\
rass & Richmond Agitation Sedation Scale & observation & neuro & categorical \\
hbco & Carboxyhemoglobin & observation & circulatory & \% \\
esr & Erythrocyte Sedimentation Rate & observation & infection & mm/hr \\
pt & Prothrombine Time & observation & circulatory & sec \\
adm & Patient Admission Type & demographic & None & categorical \\
hba1c & Hemoglobin A1C & observation & metabolic\_renal & \% \\
samp & Body Fluid Sampling, Detected Bacterial Growth & observation & infection & categorical \\
spo2 & Pulse Oxymetry Oxygen Saturation & observation & respiratory & \% \\
sao2 & Oxygen Saturation In Arterial Blood & observation & respiratory & \% \\
icp & Intra Cranial Pressure & observation & neuro & mmHg \\
cout & Cardiac Output & observation & circulatory & l/min \\
mpap & Mean Pulmonal Arterial Pressure & observation & circulatory & mmHg \\
spap & Systolic Pulmonal Arterial Pressure & observation & circulatory & mmHg \\
dpap & Diastolic Pulmonal Arterial Pressure & observation & circulatory & mmHg \\
cvp & Central Venous Pressure & observation & circulatory & mmHg \\
svo2 & Mixed Venous Oxygenation & observation & circulatory & \% \\
pcwp & Pulmonary Capillary Wedge Pressure & observation & circulatory & mmHg \\
peep & Positive End Expiratory Pressure - Mechanical Ventilation & observation & respiratory & cmH2O \\
peak & Peak Pressure - Mechanical Ventilation & observation & respiratory & cmH2O \\
plateau & Plateau Pressure - Mechanical Ventilation & observation & respiratory & cmH2O \\
ps & Pressure Support - Mechanical Ventilation & observation & respiratory & cmH2O \\
tv & Tidal Volume & observation & respiratory & ml \\
airway & Type Of Airway Ventilation & observation & respiratory & categorical \\
supp\_o2\_vent & Supplemental Oxygen From Ventilator & treatment & respiratory & \% \\
ygt & Gamma GT & observation & gastrointestinal & U/L \\
amm & Ammoniak & observation & gastrointestinal & mmol/L \\
amyl & Amylase & observation & gastrointestinal & U/L \\
lip & Lipase & observation & gastrointestinal & U/L \\
ufilt & Ultrafiltration On Continuous RRT & treatment & metabolic\_renal & ml \\
ufilt\_ind & Ultrafiltration On Continuous RRT Indicator & treatment & metabolic\_renal & indicator \\
dobu & Dobutamine & treatment & circulatory & mcg/min \\
levo & Levosimendan & treatment & circulatory & mcg/min \\
norepi & Norepinephrine & treatment & circulatory & mcg/min \\
epi & Epinephrine & treatment & circulatory & mcg/min \\
milrin & Milrinone & treatment & circulatory & mcg/min \\
teophyllin & Theophylline & treatment & circulatory & mg/min \\
dopa & Dopamine & treatment & circulatory & mcg/min \\
adh & Vasopressin & treatment & circulatory & U/min \\
hep & Heparin & treatment & circulatory & U/h \\
prop & Propofol & treatment & neuro & mcg/min \\
benzdia & Benzodiacepine & treatment & neuro & mg/h \\
sed & Other Sedatives & treatment & neuro & indicator \\
op\_pain & Opiate Painkiller & treatment & neuro & indicator \\
nonop\_pain & Non-Opioid Analgesic & treatment & neuro & indicator \\
paral & Paralytic & treatment & neuro & indicator \\
abx & Antibotics & treatment & infection & indicator \\
loop\_diur & Loop Diuretic & treatment & metabolic\_renal & mg/h \\
ins\_ind & Insulin & treatment & None & indicator \\
fluid & Fluid Administration & treatment & None & indicator \\
inf\_rbc & Packed Red Blood Cells & treatment & None & indicator \\
ffp & Fresh Frozen Plasma & treatment & None & indicator \\
plat & Platelets & treatment & None & indicator \\
inf\_alb & Albumin Infusion & treatment & None & indicator \\
anti\_delir & Anti Deliriant & treatment & neuro & indicator \\
oth\_diur & Other Diuretics & treatment & metabolic\_renal & indicator \\
anti\_coag & Other Anticoagulants & treatment & circulatory & indicator \\
vasod & Antihypertensive And Vasodilators & treatment & circulatory & indicator \\
anti\_arrhythm & Antiarrhythmic & treatment & circulatory & indicator \\
dobu\_ind & Dobutamine Indicator & treatment & circulatory & indicator \\
levo\_ind & Levosimendan Indicator & treatment & circulatory & indicator \\
norepi\_ind & Norepinephrine Indicator & treatment & circulatory & indicator \\
epi\_ind & Epinephrine & treatment & circulatory & indicator \\
milrin\_ind & Milrinone Indicator & treatment & circulatory & indicator \\
teophyllin\_ind & Theophylline Indicator & treatment & circulatory & indicator \\
dopa\_ind & Dopamine Indicator & treatment & circulatory & indicator \\
adh\_ind & Vasopressin Indicator & treatment & circulatory & indicator \\
hep\_ind & Heparin Indicator & treatment & circulatory & indicator \\
prop\_ind & Propofol Indicator & treatment & circulatory & indicator \\
benzdia\_ind & Benzodiacepine Indicator & treatment & circulatory & indicator \\
loop\_diur\_ind & Loop Diuretics Indicator & treatment & circulatory & indicator \\
dobu\_ind & Dobutamine Indicator & treatment & circulatory & indicator \\
levo\_ind & Levosimendan Indicator & treatment & circulatory & indicator \\
norepi\_ind & Norepinephrine Indicator & treatment & circulatory & indicator \\
epi\_ind & Epinephrine Indicator & treatment & circulatory & indicator \\
milrin\_ind & Milrinone Indicator & treatment & circulatory & indicator \\
teophyllin\_ind & Theophylline Indicator & treatment & circulatory & indicator \\
dopa\_ind & Dopamine Indicator & treatment & circulatory & indicator \\
adh\_ind & Vasopressin Indicator & treatment & circulatory & indicator \\
hep\_ind & Heparin Indicator & treatment & circulatory & indicator \\
prop\_ind & Propofol Indicator & treatment & neuro & indicator \\
benzdia\_ind & Benzodiacepine Indicator & treatment & neuro & indicator \\
loop\_diur\_ind & Loop Diuretic Indicator & treatment & metabolic\_renal & indicator \\* \bottomrule
\caption{Concept reference}
\label{tab:variables}\\
\end{longtable}
\normalsize 
\subsection{Pre-processing}

\subsubsection{Task Annotations}

We define sample labels by annotating the time series following the clinical definitions used by prior work for each specific task. As such, these labels constitute proxy labels derived and computed from data given a clinical definition to diagnose a patient state for a certain condition. Most importantly, we annotate a positive and a negative case only if there's enough evidence in favor of either. As such, a label is only computed if the source data, conditioned on a task-specific imputation scheme proposed by prior work, provides all required inputs to compute the score or state annotation of the clinical task definition at a certain time step. Based on these cases, we define early event prediction labels (e.g., respiratory failure) that are then used for online classification of the future state of the patient.

Early event prediction (EEP) label for a given time step is computed as follows: (1) a detection (positive EEP label) is marked if any time-point in the future within the horizon is annotated as the patient is in a failure state; (2) a negative EEP label (a stable patient without any upcoming failure state) is annotated only if there is no failure state annotation and there is at least one confirmed stable state within the horizon; (3) if there is no data confirmed evidence for either the patient being in failure or being stable within the horizon, no EEP label is assigned and no training and evaluation is performed for that specific time step. We use task-specific and clinically relevant prediction horizons from existing literature (8 hours for circulatory failure~\citep{hyland2020early}, 24 hours for decompensation~\citep{harutyunyan2019multitask} and respiratory failure~\cite{huser2024medrxiv_respiratory}, and 48 hours for kidney failure~\cite{lyu2024empirical}).

\subsubsection{Data Scaling}
Data is scaled depending on its type:
\begin{itemize}
    \item continuous observations are standardized (i.e. centered and scaled to unit variance),
    \item categorical observations are one-hot encoded and each variable has a dedicated class to encode missing information,
    \item continuous treatments are quantile-transformed and mapped to the $[0, 1]$ range such that a $0$ represents \emph{no medication given},
    \item treatment indicators are binary encoded using $\{0, 1\}$.
\end{itemize}

\subsubsection{Imputation}

Gridded time-step data as inputs for model training are forward-filled indefinitely for all observation variables. The remaining missing values are then imputed with $0$ for continuous variables, which corresponds to a population mean imputation after considering standard scaling before the imputation stage. The remaining categorical entries are imputed with a value corresponding to the dedicated class that encodes missing information for each categorical variable.

Any treatment variable is excluded from forward-filling operations and missing data points are strictly filled using $0$, which given the previously introduced scaling and encoding scheme always corresponds to no treatment being applied.

\subsubsection{Feature Extraction}
\label{sec:app_features}

We build on the feature set proposed by~\citet{soenksen2022multimodalai} to process the MIMIC-IV~\citep{MIMIC-IV} dataset. To improve performance we then further expand this set of features and select specific features for each variable type. For each time step, each feature is computed over three history sizes of 8, 24, and 72 hours:

\begin{itemize}
    \item For continuous observations and continuous treatment variables, we compute:
        \begin{itemize}
            \item mean on raw and imputed data,
            \item standard deviation on raw data,
            \item slope of a linear fit on the raw data and imputed data,
            \item mean absolute change over imputed data,
            \item fraction of non-missing data points,
            \item quantiles: 0\% (Min.), 10\%, 50\%, 90\%, 100\% (Max.).
        \end{itemize}
    \item For categorical variables, we compute the mode, number of missing points, and a binary indicator of whether there are any missing points at all.
    \item For treatment indicators we compute the number of points with treatment and a binary indicator whether any treatment was applied.
\end{itemize}

\section{Training details}

We evaluate the performance of 7 model architectures. For each, we find the best set of hyperparameters using grid search with a set of approximately 12 points per model. For deep learning architectures, we focus on hidden dimensions, number of layers, and architecture-specific parameters. For LightGBM~\citep{ke2017lightgbm} we choose a strong starting point based on hyperparameters reported by~\citet{yeche2021, hyland2020early} and then further tune: \texttt{colsample\_bytree}, \texttt{subsample}, \texttt{}{num\_leaves}, \texttt{min\_child\_samples}, and \texttt{subsample\_for\_bin}. For linear models trained using \texttt{glum}~\citep{glum2020quantco} we optimize regularization parameters. For each set of hyperparameters, model performance is evaluated as an average across three seeds. 

Single-center experiments involve training on every dataset and evaluating on every other dataset for each task, resulting in 30 training runs per architecture. Multi-center experiments involve training on all datasets except one in a leave-one-out fashion, also resulting in 30 runs, but with larger training sets. For disposition prediction experiments training is performed once for each model, as it is not a transfer study.

Overall, approximately $7 \cdot 12 \cdot (30 + 30 + 1) = 5124$ runs were performed. We use one to four top-of-the-line Nvidia H200 GPUs with up to 100GB of GPU memory for each run,  depending on the task, architecture, and training set size. Multi-card training is done using \texttt{pytorch-lightning} \citep{Falcon_PyTorch_Lightning_2019}. Each compute server, an Nvidia Grace Hopper GH200 Superchip server, is equipped with up to 400GB of main memory, an ARM CPU with up to 288 cores, and has 4 GPUs. Experiments were run on a cluster infrastructure providing many servers with the aforementioned specifications. Each experiment shown in the paper was run on a single node (server).

\begin{figure}[htbp]
    \centering
\begin{subfigure}[b]{0.45\textwidth}
        \centering
        \includegraphics[width=0.95\linewidth]{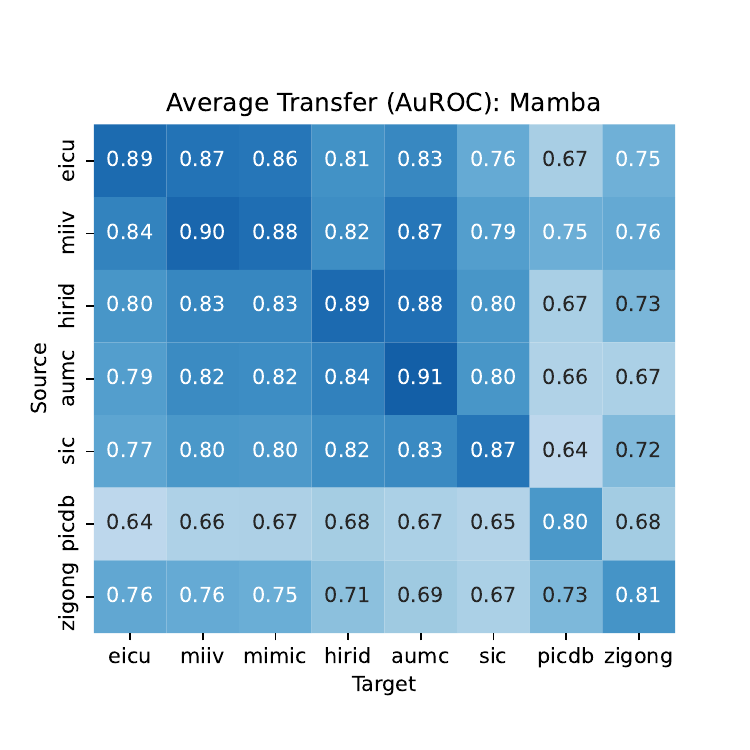}
        \caption{Mamba task-average}
        \label{fig:mamba-avg-transfer}
    \end{subfigure}
    \hfill
\begin{subfigure}[b]{0.45\textwidth}
        \centering
        \includegraphics[width=0.95\linewidth]{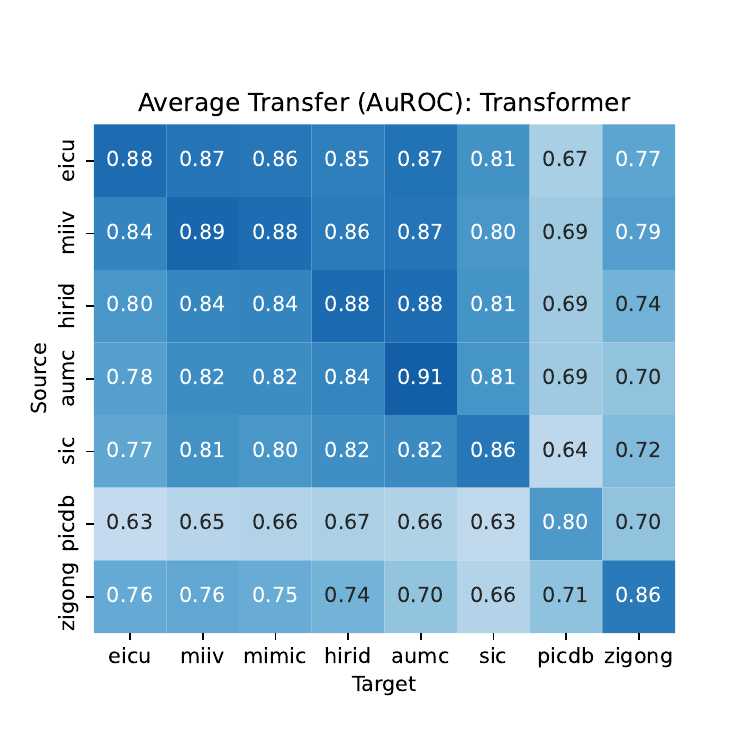}
        \caption{Transformer task-average}
        \label{fig:transformer-avg-transfer}
    \end{subfigure}
    \caption{Single-center transfer performance heatmaps (AUROC).}
    \label{fig:transformer-mamba-avg-transfer}
\end{figure}

\section{Additional results}
\label{sec:app_add_results}

\subsection{Transfer Heatmaps}

\cref{fig:transformer-mamba-avg-transfer} shows additional single-center transfer results for Mamba~\citep{dao2024transformersssmsgeneralizedmodels} and Transformer~\citep{vaswani2017attention}.

\subsection{Fine-tuning Study}

\begin{figure}[htbp]
    \centering
\begin{subfigure}[b]{0.45\textwidth}
        \centering
        \includegraphics[width=0.9\linewidth]{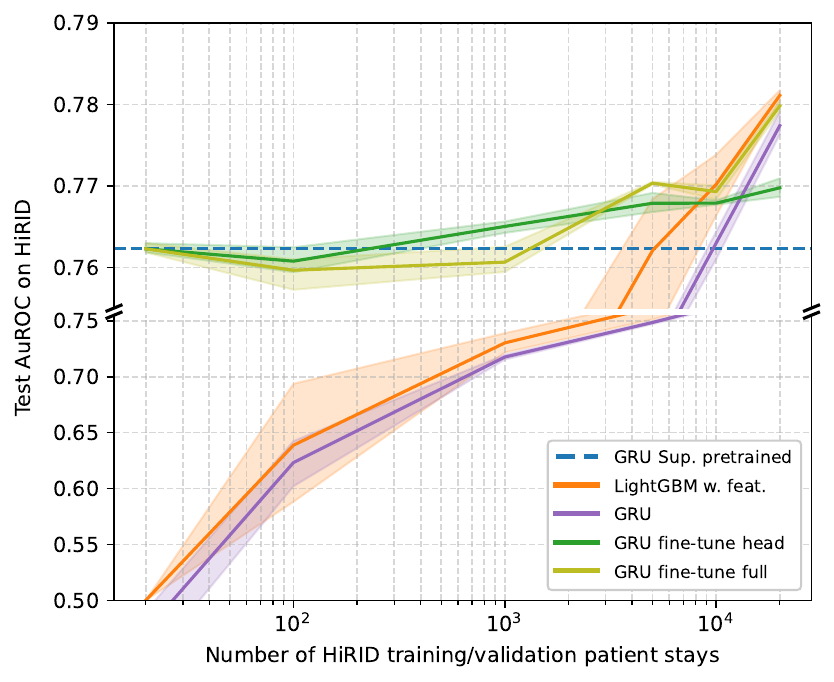}
        \caption{Resp. 24h (AUROC)}
        \label{fig:fine-tuning-hirid-auroc-resp}
    \end{subfigure}
    \hfill
\begin{subfigure}[b]{0.45\textwidth}
        \centering
        \includegraphics[width=0.9\linewidth]{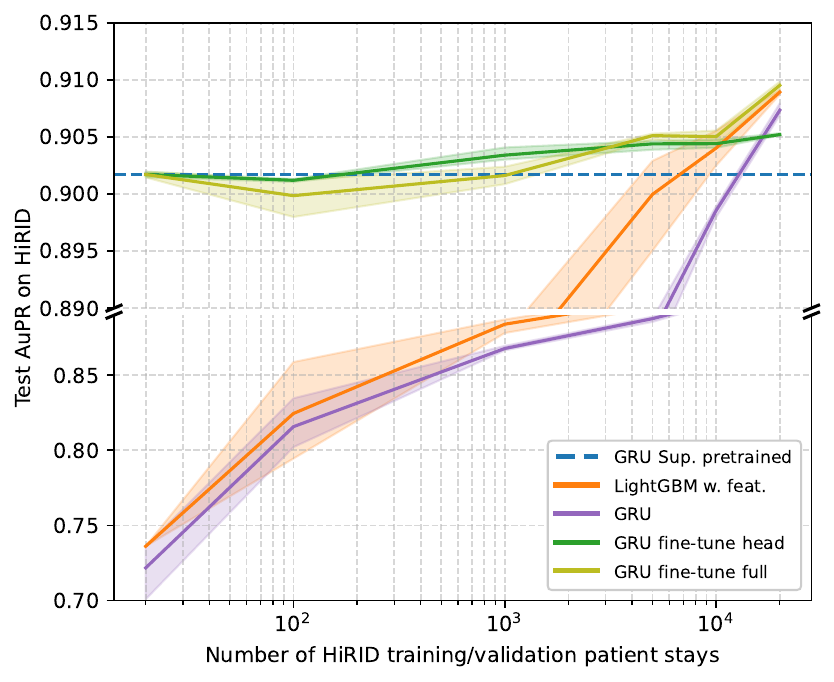}
        \caption{Resp. 24h (AUPRC)}
        \label{fig:fine-tuning-hirid-aupr-resp}
    \end{subfigure}

    \begin{subfigure}[b]{0.45\textwidth}
        \centering
        \includegraphics[width=0.9\linewidth]{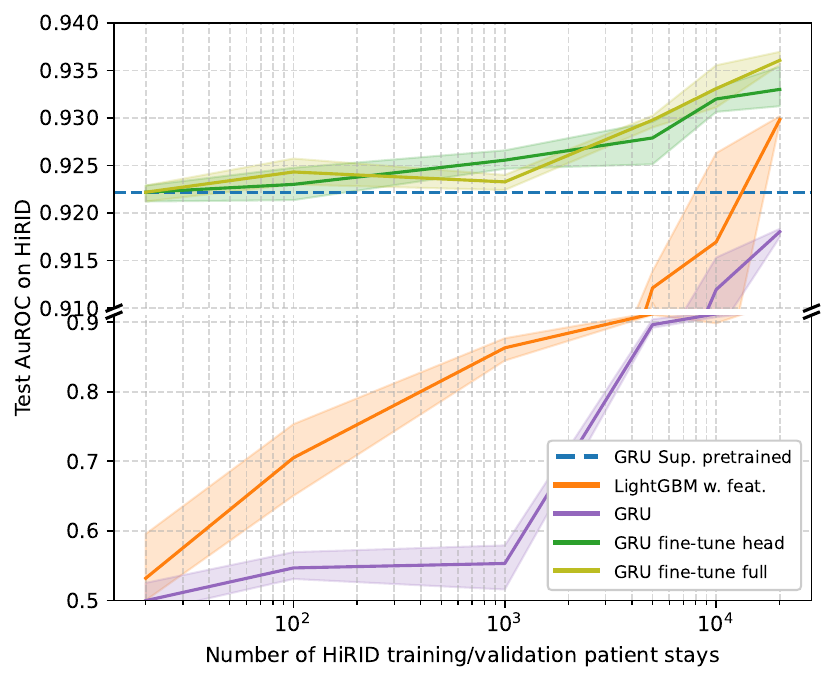}
        \caption{Kidney 24h (AUROC)}
        \label{fig:fine-tuning-hirid-auroc-kidney}
    \end{subfigure}
    \hfill
\begin{subfigure}[b]{0.45\textwidth}
        \centering
        \includegraphics[width=0.9\linewidth]{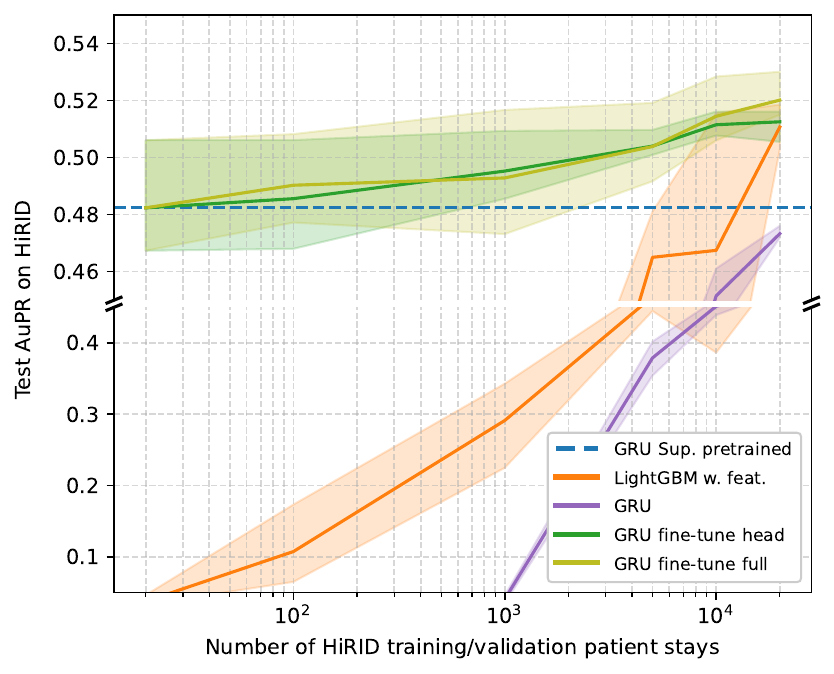}
        \caption{Kidney 24h (AUPRC)}
        \label{fig:fine-tuning-hirid-aupr-kidney}
    \end{subfigure}
    
    \caption{Supervised fine-tuning study performed on HiRID for respiratory failure (\cref{fig:fine-tuning-hirid-aupr-resp,fig:fine-tuning-hirid-auroc-kidney}) and kidney failure (\cref{fig:fine-tuning-hirid-auroc-kidney,fig:fine-tuning-hirid-aupr-kidney}) by progressively increasing the number of patients shown during training or fine-tuning. \emph{GRU} and \emph{LGBM w. feat.} are trained from scratch using HiRID data only. \emph{GRU pretrained} is trained on all data excluding HiRID patients. \emph{GRU fine-tuned (head/full)} initialize the network with \emph{GRU pretrained} and fine-tune the full network or only the single linear logit head.}
    \label{fig:fine-tuning-study-resp-kidney}
\end{figure}

In~\cref{fig:fine-tuning-study-resp-kidney} we show further fine-tuning study results for respiratory failure and kidney failure predictions, which confirm the trend already highlighted and discussed in~\cref{fig:fine-tuning-study}. The performed data harmonization work is highly valuable for small to medium-sized hospitals, which only have limited amounts of training patients available and can thus significantly benefit from pretraining on data from other hospitals.

\subsection{Benchmarking Results for AUPRC}

\cref{tab:benchmark-id-aupr} shows the in-distribution benchmarking results using the area under the precision-recall curve metric and corresponds to the AUROC results shown in~\cref{tab:benchmark-id}.

\cref{tab:benchmark-ood-aupr} shows the out-of-distribution benchmarking results using the area under the precision-recall curve metric and corresponds to the AUROC results shown in~\cref{tab:benchmark-ood}.

\clearpage{}\begin{table*}[tb]
\centering
\resizebox{\textwidth}{!}{\begin{tabular}{l l c c c c c c c c c c c c c c}
    \toprule
    \bf Dataset & \bf Task & \multicolumn{7}{c}{\textbf{Single-Center}} & \multicolumn{7}{c}{\textbf{Multi-Center}} \\
    \cmidrule(lr){3-9} \cmidrule(lr){10-16} 
     & & \rot{LR (Last Meas.)} & \rot{LGBM (Last Meas.)} & \rot{LGBM (Feat.)} & \rot{GRU} & \rot{Transformer} & \rot{Mamba}  & \rot{xLSTM} & \rot{LR (Last Meas.)} & \rot{LGBM (Last Meas.)} & \rot{LGBM (Feat.)} & \rot{GRU} & \rot{Transformer} & \rot{Mamba}  & \rot{xLSTM} \\

\midrule
    \multirow{4}{*}{\norot{MIMIC-IV}} & Dec. 24h &  44.7 & 53.5 & \textbf{62.4} & 57.4 & 55.6 & 56.4 & 56.4 & 39.2 & 49.8 & 56.4 & 52.9 & 49.5 & 53.1 & 53.5 \\
    & Circ. 8h  & 59.7 & 66.8 & \textbf{68.2} & 65.7 & 65.4 & 65.3 & 66.2 & 55.7 & 63.8 & 66.0 & 63.7 & 63.2 & 63.9 & 64.8  \\
    & Resp. 24h & 75.4 & 79.6 & \textbf{81.2} & 79.6 & 79.2 & 79.3 & 79.4 & 73.6 & 78.5 & 80.0 & 79.4 & 78.7 & 79.4 & 79.4  \\
    & Kidn. 48h & 40.8 & 48.6 & \textbf{51.8} & 46.9 & 45.6 & 46.0 & 47.0 & 39.5 & 47.2 & 50.8 & 47.6 & 45.9 & 48.3 & 48.6  \\

\midrule
    \multirow{4}{*}{\norot{eICU}} & Dec. 24h &  33.1 & 37.9 & \textbf{51.4} & 41.2 & 40.9 & 40.8 & 40.9 & 30.0 & 36.8 & 41.8 & 39.5 & 38.0 & 38.9 & 39.5  \\
    & Circ. 8h  &  56.8 & 64.7 & \textbf{65.3} & 63.8 & 63.2 & 62.8 & 63.5 & 54.2 & 61.7 & 63.4 & 60.9 & 60.6 & 61.4 & 62.2  \\
    & Resp. 24h &  72.6 & 78.2 & \textbf{80.0} & 79.1 & 78.3 & 78.4 & 78.3 & 71.3 & 77.7 & 78.7 & 78.0 & 77.1 & 77.9 & 78.2  \\
    & Kidn. 48h &  33.6 & 42.5 & \textbf{47.3} & 43.3 & 42.5 & 43.1 & 43.8 & 32.7 & 40.7 & 44.5 & 41.2 & 40.7 & 42.4 & 42.9  \\

\midrule
    \multirow{4}{*}{\norot{HiRID}} & Dec. 24h &  43.3 & 50.1 & 52.9 & 54.1 & 51.9 & 52.7 & 53.4 & 42.1 & 51.3 & \textbf{55.7} & 53.4 & 51.8 & 52.5 & 52.7  \\
    & Circ. 8h  &  52.5 & 55.4 & 57.3 & 57.6 & 56.7 & 56.9 & 57.6 & 51.8 & 57.0 & 59.0 & \textbf{59.2} & 58.3 & 57.8 & 58.6  \\
    & Resp. 24h &  88.7 & 90.1 & 90.9 & 90.3 & 90.1 & 89.9 & 90.0 & 88.3 & 90.4 & \textbf{91.0} & 90.9 & 90.6 & 90.6 & 90.6  \\
    & Kidn. 48h &  44.4 & 52.6 & 54.6 & 50.1 & 48.8 & 47.9 & 48.7 & 41.4 & 53.8 & \textbf{57.9} & 50.7 & 52.8 & 49.6 & 51.4  \\

\midrule
    \multirow{4}{*}{\norot{UMCdb}} & Dec. 24h &  36.0 & 42.2 & \textbf{55.7} & 47.1 & 44.0 & 43.8 & 43.6 & 35.4 & 43.7 & 54.4 & 50.5 & 48.5 & 48.5 & 48.1  \\
    & Circ. 8h  &   85.5 & 89.8 & \textbf{91.7} & 90.8 & 90.3 & 90.5 & 90.8 & 82.2 & 87.6 & 91.4 & 91.4 & 91.4 & 91.4 & 91.7  \\
    & Resp. 24h &  85.3 & 86.9 & \textbf{88.1} & 87.5 & 87.0 & 86.9 & 86.8 & 84.6 & 87.0 & 88.1 & 87.7 & 87.5 & 87.3 & 87.1  \\
    & Kidn. 48h &  50.8 & 56.5 & 58.9 & 56.0 & 53.9 & 53.6 & 54.2 & 49.1 & 59.1 & \textbf{63.3} & 56.6 & 58.5 & 55.2 & 57.2  \\
    
\midrule
    \multirow{4}{*}{\norot{SICdb}} & Dec. 24h &  31.0 & 31.7 & 32.6 & 33.4 & 34.5 & 32.3 & 32.9 & 28.4 & 35.6 & 39.7 & 38.5 & 38.5 & 38.1 & \textbf{40.4}  \\
    & Circ. 8h  &   49.1 & 53.4 & 56.2 & 55.1 & 54.0 & 54.3 & 54.1 & 47.8 & 53.9 & 56.6 & 56.6 & 55.8 & 55.8 & \textbf{57.0}  \\
    & Resp. 24h &   85.6 & 88.1 & 88.6 & 88.0 & 87.6 & 87.4 & 87.3 & 85.5 & 88.3 & \textbf{89.0} & 88.6 & 88.2 & 88.5 & 88.4  \\
    & Kidn. 48h &   31.5 & 33.9 & 42.6 & 35.9 & 31.4 & 32.4 & 33.1 & 29.1 & 39.5 & \textbf{45.8} & 37.0 & 37.9 & 35.8 & 36.2  \\

\midrule
    \multirow{4}{*}{\norot{PICdb}} & Dec. 24h &  15.2 & 12.0 & 16.5 & 16.6 & 16.5 & 16.2 & 16.4 & 6.9 & 16.0 & \textbf{19.1} & 13.7 & 12.0 & 13.5 & 13.2  \\
    & Circ. 8h  &  92.9 & 93.9 & \textbf{96.3} & 95.1 & 95.2 & 95.0 & 95.7 & 87.0 & 91.1 & 90.9 & 90.8 & 91.1 & 92.2 & 92.8  \\
    & Resp. 24h &  8.2 & 8.2 & \textbf{11.3} & 10.6 & 10.4 & 11.1 & 9.9 & 4.6 & 3.6 & 3.7 & 4.8 & 5.5 & 5.5 & 7.2  \\
    & Kidn. 48h &  5.1 & 9.2 & \textbf{11.8} & 3.2 & 3.7 & 3.8 & 3.9 & 6.6 & 7.6 & 11.3 & 9.3 & 8.4 & 7.3 & 7.1  \\

\midrule
    \multirow{2}{*}{\norot{Zigong}} & Dec. 24h &  22.3 & 30.3 & \textbf{54.0} & 41.4 & 37.8 & 29.9 & 32.6 & 20.3 & 20.4 & 28.3 & 23.2 & 24.5 & 22.5 & 25.5  \\
    & Circ. 8h  & 98.0 & 98.6 & \textbf{98.8} & 98.0 & 97.5 & 97.0 & 94.4 & 98.2 & 98.3 & 98.2 & 97.6 & 97.2 & 97.3 & 97.5  \\

    \bottomrule
    \end{tabular}
}
    \caption{Benchmarking results in-distribution (AUPRC). Bold is best in each row. Multi-center trains on all datasets together and provides the individual test performances. All results show the mean over three different random initialization, except for LR models that are trained using convex optimization. }

    \label{tab:benchmark-id-aupr}
\end{table*}\clearpage{}
\clearpage{}
\begin{table*}[tb]
\centering
\resizebox{\textwidth}{!}{\begin{tabular}{l l c c c | c c c c c c c}
    \toprule
    \bf Dataset & \bf Task & \multicolumn{3}{c}{\textbf{Single-Center}} & \multicolumn{7}{c}{\textbf{Multi-Center hold-out}} \\
    \cmidrule(lr){3-5} \cmidrule(lr){6-12} 
     & & AUPRC & Best Model & Src. Data & 
     \rot{LR (Last Meas.)} & \rot{LGBM (Last Meas.)} & \rot{LGBM (Feat.)} & \rot{GRU} & \rot{Transformer} & \rot{Mamba}  & \rot{xLSTM} \\

\midrule
    \multirow{4}{*}{\norot{MIMIC-IV}} & Dec. 24h & \textbf{55.5} & LGBM (Feat.) & eICU &  36.9 & 46.0 & 54.0 & 47.6 & 43.8 & 48.7 & 46.8  \\
    & Circ. 8h  & \textbf{62.9} & LGBM (Feat.) & eICU &  52.2 & 56.0 & \textbf{62.9} & 59.0 & 58.1 & 57.4 & 57.9  \\
    & Resp. 24h & \textbf{78.8} & LGBM (Feat.) & eICU &  72.1 & 76.3 & 78.4 & 76.9 & 76.5 & 76.9 & 76.6  \\
    & Kidn. 48h & \textbf{51.7} & LGBM (Feat.) & eICU &  38.9 & 45.9 & 48.9 & 44.1 & 43.1 & 44.9 & 42.5  \\

\midrule
    \multirow{4}{*}{\norot{eICU}} & Dec. 24h & \textbf{38.2} & LGBM (Feat.) & MIMIC-IV &  27.1 & 34.3 & 37.0 & 34.3 & 35.0 & 33.3 & 33.7  \\
    & Circ. 8h  & \textbf{62.2} & LGBM (Feat.) & MIMIC-IV &  52.9 & 54.8 & 61.6 & 56.4 & 55.6 & 57.5 & 57.5  \\
    & Resp. 24h & \textbf{77.5} & LGBM (Feat.) & MIMIC-IV &  69.5 & 75.7 & 76.8 & 74.4 & 73.3 & 74.9 & 74.2  \\
    & Kidn. 48h & \textbf{43.8} & LGBM (Feat.) & MIMIC-IV &  31.9 & 39.1 & 42.3 & 34.3 & 34.8 & 35.3 & 34.7  \\

\midrule
    \multirow{4}{*}{\norot{HiRID}} & Dec. 24h & 39.3 & LGBM (Feat.) & UMCdb &  38.4 & 42.2 & 42.5 & 43.1 & \textbf{45.4} & 43.9 & 42.8  \\
    & Circ. 8h  & 50.6 & LGBM (Feat.) & SICdb    &  49.8 & 52.8 & \textbf{53.5} & 51.7 & 52.0 & 48.3 & 49.3  \\
    & Resp. 24h & 89.7 & LGBM (Feat.) & MIMIC-IV &  87.7 & 89.2 & \textbf{90.0} & 90.0 & 89.9 & 89.4 & 89.4  \\
    & Kidn. 48h & \textbf{52.8} & LGBM (Feat.) & MIMIC-IV &  39.2 & 51.4 & \textbf{55.7} & 48.3 & 47.7 & 44.0 & 46.7  \\

\midrule
    \multirow{4}{*}{\norot{UMCdb}} & Dec. 24h & 38.2 & LGBM (Feat.) & HiRID &  34.3 & 38.6 & \textbf{41.9} & 41.6 & 40.9 & 39.5 & 39.0  \\
    & Circ. 8h  & \textbf{85.5} & GRU & HiRID &  80.0 & 81.6 & 83.6 & 85.4 & 83.9 & 84.1 & 84.9  \\
    & Resp. 24h & \textbf{86.7} & LGBM (Feat.) & eICU &  83.8 & 85.5 & 86.6 & 86.3 & 85.1 & 84.4 & 84.6  \\
    & Kidn. 48h & 59.4 & LGBM (Last Meas.) & MIMIC-IV &  46.7 & 56.9 & \textbf{60.6} & 52.6 & 49.9 & 50.9 & 45.4  \\
    
\midrule
    \multirow{4}{*}{\norot{SICdb}} & Dec. 24h & 28.6 & LGBM (Feat.) & eICU &  24.1 & \textbf{31.0} & 30.8 & 29.4 & 28.0 & 28.3 & 28.9  \\
    & Circ. 8h  & 46.9 & LGBM (Feat.) & HiRID &  46.1 & 47.9 & 48.1 & 46.2 & \textbf{48.7} & 45.8 & 46.3  \\
    & Resp. 24h & 87.6 & LGBM (Feat.) & eICU &  84.8 & 87.0 & \textbf{87.7} & 86.9 & 86.6 & 86.6 & 86.7  \\
    & Kidn. 48h & 37.0 & LGBM (Feat.) & UMCdb & 25.4 & 36.0 & \textbf{41.6} & 31.9 & 32.1 & 32.1 & 30.9  \\

\midrule
    \multirow{4}{*}{\norot{PICdb}} & Dec. 24h & 7.9 & LGBM (Feat.) & HiRID &  5.9 & 7.3 & \textbf{8.3} & 6.5 & 4.5 & 5.1 & 2.7  \\
    & Circ. 8h  & \textbf{89.4} & GRU & MIMIC-IV &  86.7 & 86.9 & 85.6 & 86.8 & 86.1 & 85.6 & 85.4  \\
    & Resp. 24h & \textbf{7.7} & GRU & HiRID &  5.1 & 4.1 & 7.3 & 5.8 & 5.8 & 5.2 & 4.8  \\
    & Kidn. 48h & 8.1 & LGBM (Last Meas.) & HiRID &  6.6 & 7.3 & 7.5 & 7.7 & 8.0 & 7.9 & \textbf{8.8}  \\

\midrule
    \multirow{2}{*}{\norot{Zigong}} & Dec. 24h & \textbf{22.2} & LGBM (Feat.) & UMCdb &  20.8 & 19.1 & 20.3 & 17.2 & 18.1 & 18.9 & 18.8  \\
    & Circ. 8h  & \textbf{98.6} & LGBM (Last Meas.) & MIMIC-IV & 98.2 & 97.9 & 98.2 & 97.9 & 97.7 & 97.0 & 97.1  \\

    \bottomrule
    \end{tabular}
}     
    \caption{Benchmarking results out-of-distribution (AUPRC). Bold is best in each row (separately for single-cente and multi-center). Single-center results are an argmax over training datasets while testing on a hold-out dataset. Multi-center models are trained on all but the test dataset. All results show the mean over three different random initialization, except for LR models that are trained using convex optimization.}
\label{tab:benchmark-ood-aupr}
\end{table*}\clearpage{}

\section{Impact and limitations}

\emph{Impact.} This work advances ML research for healthcare by enhancing models for early event prediction of adverse medical conditions. This research could in the future lead to improved care for patients in emergency units. This research incorporates data from multiple continents, making ML research in critical care time series more accessible and fair. Potential harmful impacts may include compromised patient safety. Investigation of the models to ensure their fairness, robustness, and privacy is an open topic for future works.

\emph{Limitations.} This work has considered online early event prediction tasks as they are clinically relevant and typically harder than the alternatives. Other tasks can be considered (e.g., prediction of mortality, length of stay, sepsis)~\citep{moor2023predicting,hirid}. We limit hyperparameter search to approximately 12 points per model due to the huge computational burden of running the benchmark experiments (multiple seeds, multiple datasets, multiple tasks).

\end{document}